\documentclass[runningheads]{llncs}

\usepackage{eccv}

\usepackage{eccvabbrv}

\usepackage{graphicx}
\usepackage{booktabs}

\usepackage[accsupp]{axessibility}  

\usepackage{hyperref}

\usepackage{booktabs}

\usepackage{multirow}
\usepackage{adjustbox}
\usepackage{array}
\usepackage{wrapfig,lipsum}
\usepackage[font={footnotesize}]{caption}
\usepackage{makecell}

\usepackage{adjustbox}
\usepackage{graphicx}
\usepackage{fix-cm}
\makeatletter
\def\thickhline{%
  \noalign{\ifnum0=`}\fi\hrule \@height \thickarrayrulewidth \futurelet
   \reserved@a\@xthickhline}
\def\@xthickhline{\ifx\reserved@a\thickhline
               \vskip\doublerulesep
               \vskip-\thickarrayrulewidth
             \fi
      \ifnum0=`{\fi}}
\makeatother
\newlength{\thickarrayrulewidth}
\usepackage{makecell}

\usepackage{adjustbox}
\usepackage{graphicx}
\usepackage{fix-cm}
\makeatletter
\def\thickhline{%
  \noalign{\ifnum0=`}\fi\hrule \@height \thickarrayrulewidth \futurelet
   \reserved@a\@xthickhline}
   
\def\@xthickhline{\ifx\reserved@a\thickhline
               \vskip\doublerulesep
               \vskip-\thickarrayrulewidth
             \fi
      \ifnum0=`{\fi}}
\makeatother
\usepackage[dvipsnames]{xcolor}
\usepackage{pifont}
\usepackage{graphicx}
\usepackage{colortbl}
\usepackage{arydshln}
\usepackage{tikz}
\definecolor{Gray}{gray}{0.8}
\definecolor{LG}{gray}{.92}

\newcommand{\gc}{\cellcolor{gray!20}}

\newcommand{\gr}{\rowcolor{gray!20}}

\newcommand{\gcc}{\cellcolor{gray!40}}

\newcommand{\cmark}{\textcolor{green}{\ding{51}}}

\usepackage{multirow}
\usepackage{makecell}
\usepackage{subcaption}
\usepackage{setspace}
\usepackage{bm}
\usepackage{tikz}

\usepackage{enumitem}
\usepackage{lipsum}

\usepackage{orcidlink}

\begin{document}









\title{GoStop: Reinforcement Learning for \\ Adaptive Temporal Aggregation in \\ Event-Based Feature Tracking}


\titlerunning{GoStop}

\author{Youngho Kim$^{*}$\orcidlink{0009-0005-4024-5031} \and
Hoonhee Cho$^{*}$\orcidlink{0000-0003-0896-6793} \and
Jae-Young Kang\orcidlink{0009-0002-9537-3813} \and
Kuk-Jin Yoon\orcidlink{0000-0002-1634-2756}}
\authorrunning{Y.~Kim et al.}
\institute{KAIST 
\\
\email{\{kmax2001, gnsgnsgml, kjyoon, mickeykang\}@kaist.ac.kr}\\
}
\maketitle

\begin{abstract}
Feature tracking plays a fundamental role in understanding scene motion and supports various downstream tasks. Event cameras, with their high temporal resolution and asynchronous sensing, enable low-latency and motion-robust perception, making them well-suited for feature tracking under fast and non-linear motion. However, existing event-based feature tracking methods rely on fixed heuristic rules based on hand-tuning for event accumulation. Such strategies fail to adapt to diverse motion dynamics, leading to degraded performance under abrupt motion changes or low-motion scenarios. In this paper, we model event accumulation as a sequential decision-making problem and introduce reinforcement learning (RL) framework to adaptively control the accumulation process for online event-based feature tracking. Our approach trains a RL agent that decides whether to continue accumulating events or to perform tracking inference based on motion cues. The proposed adaptive temporal agent enables dynamic adaptation to varying motion patterns without relying on hand-crafted rules. Furthermore, we introduce a Dynamic Event-based Tracking (DEFT) dataset with dynamic motion distributions to evaluate the robustness of the feature tracking. Extensive experiments demonstrate that integrating our plug-and-play framework to existing feature tracking methods consistently outperforms heuristic-based approaches, improving robustness under dynamic motion while offering a better balance between tracking accuracy and efficiency. Our project codes and datasets are available at \url{https://github.com/kmax2001/GoSTOP}.
\end{abstract}    
\section{Introduction}
\label{sec:intro}

\def\thefootnote{*}\footnotetext{The first two authors contributed equally.}\def\thefootnote{\arabic{footnote}} 

Feature tracking~\cite{doersch2022tap, harley2022particle} estimates the trajectories of query points, enabling a deeper understanding of scene motion and supporting various downstream tasks~\cite{schonberger2016pixelwise, chen2024leap, vecerik2024robotap}. 
Event cameras~\cite{chakravarthi2024recent, gallego2020event}, with high dynamic range, low power consumption, and asynchronous per-pixel operation, enable low-latency sensing that is robust to motion blur and well-suited for fast ego and object motion~\cite{kang2026event6d, kim2025sharp, cho2024benchmark}, motivating event-based feature tracking~\cite{gehrig2018asynchronous, gehrig2020eklt, alzugaray2020haste}. Recent learning-based approaches~\cite{shen2025blinktrack, messikommer2023data, li20243d, messikommer2025data, hamann2025etap, han2025mate_track} leverage these properties to achieve accurate and efficient tracking, particularly under dynamic and non-linear motion, where frame-based cameras often struggle.

\begin{figure*}[t]
    \centering
    \includegraphics[width=.96\linewidth]{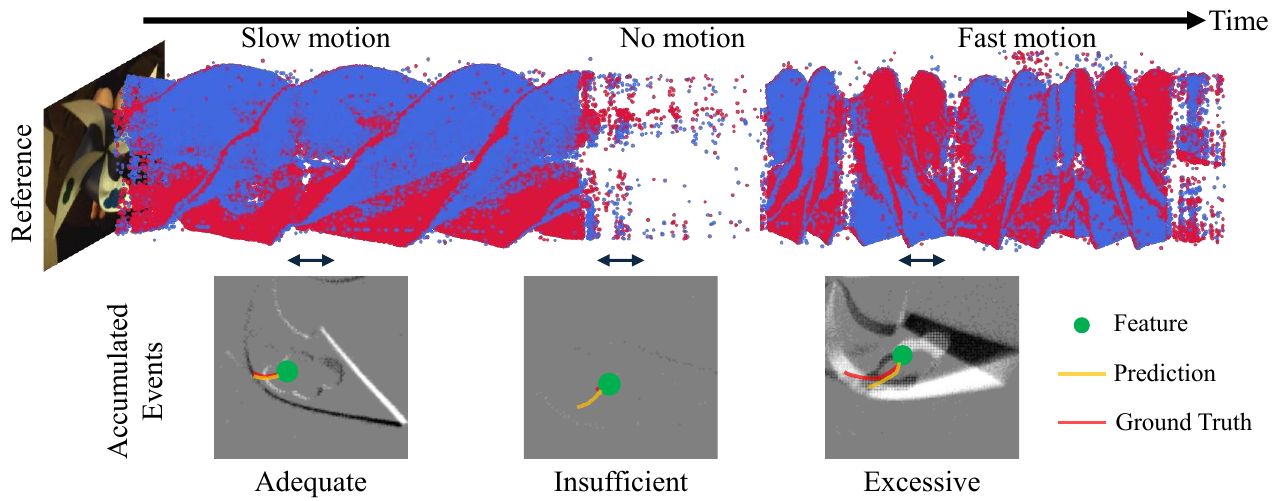}
    \vspace{-5pt}
    \caption{
    Illustration of event accumulation under a fixed temporal window across different motion states. Because the same temporal window is used regardless of motion dynamics, the accumulated events can provide insufficient cues under little motion and blurred, noisy cues under fast motion. This motivates our adaptive event accumulation strategy, which adjusts the temporal window according to the observed motion.}
    \label{fig:motivation}
    \vspace{-10pt}
\end{figure*}

However, event cameras generate a variable amount of information depending on scene dynamics, such as object motion. This temporal variability poses a significant challenge for sequential tasks such as event-based feature tracking, where the quality of each prediction directly affects subsequent predictions. To address this issue, event-based feature tracking methods~\cite{messikommer2023data, shen2025blinktrack} typically rely on manually designed rules, such as fixed-length temporal windows with hand-tuned hyperparameters.
These rules are often based on prior knowledge and tuned offline, making them difficult to generalize across diverse motion patterns and dynamic scenes. As a result, they fail to adapt to varying motion dynamics, limiting the ability to fully leverage the asynchronous nature of event cameras beyond a fixed frame rate. As illustrated in Fig.~\ref{fig:motivation}, tracking performs well when sufficient informative events are accumulated, but degrades significantly when motion suddenly diminishes or becomes rapid. 
We observe that learning-based event tracking becomes unstable when motion ceases after continuous movement and also degrades under abrupt motion changes, suggesting that a fixed accumulation interval is insufficient for robust event-based tracking.

To address this limitation, we propose to adaptively determine event accumulation in event-based feature tracking by formulating it as a sequential decision-making problem, where actions at each timestep influence future states. Deciding how long to accumulate events in a causal, online tracking setting is highly challenging due to the temporal and motion dependencies of event streams and the absence of explicit supervision for optimal accumulation. To this end, we introduce reinforcement learning (RL) to event-based feature tracking for the first time. We model the accumulation process as an environment, where an agent observes the current tracking state, such as motion cues and intermediate event statistics, and decides whether to continue accumulating events or to perform tracking inference, analogous to step-by-step decision-making. The agent is trained to optimize temporal window and maximize tracking performance through a reward function that reflects the quality of tracking outcomes, enabling the model to learn a policy that dynamically adapts accumulation strategies to diverse motion patterns while reducing reliance on hand-crafted rules and improving robustness and generalization. Compared to existing event-based feature tracking methods, our approach remains robust under varying motion distributions. By adaptively controlling the accumulation process, it gathers sufficient information for reliable inference while avoiding unnecessary accumulation, leading to a favorable trade-off between accuracy and efficiency. Furthermore, our method can be readily integrated into existing well-designed event-based tracking frameworks, making it a general and extensible solution that can be applied to future event-based feature tracking systems. The core contributions of our work can be summarized as follows:
\begin{itemize}[noitemsep, topsep=0pt]
\item We propose a RL-based event-based feature tracking framework that adaptively determines the event accumulation process in a causal and online manner.

\item We introduce a challenging dynamic event-based tracking (DEFT) dataset with diverse and rapidly changing motion distributions, including complex and non-linear motion patterns.

\end{itemize}

\section{Related Works}
\label{sec:related_works}

\noindent
\textbf{Event-based Feature Tracking.}
Frame-based feature tracking~\cite{harley2022particle, bian2023context, doersch2022tap, karaev2024cotracker, karaev2025cotracker3} is inherently limited by the camera frame rate, making it difficult to handle non-linear motion. In contrast, event cameras provide high temporal resolution, enabling event-based tracking methods to operate at high effective frame rates and better capture complex motion.
Early event-based approaches~\cite{alzugaray2018ace, chui2021event, dardelet2021event, hu2022ecdt, seok2020robust, wang2023event} relied on hand-crafted methods, which require extensive parameter tuning and often generalize poorly to new environments. Recent learning-based methods have reduced the need for such tuning. For instance, Deep-Ev-Tracker~\cite{messikommer2023data} demonstrates strong cross-dataset generalization using data-driven learning. ETAP~\cite{hamann2025etap} leverages temporal correlations across event sequences to achieve high accuracy, but incurs high computational cost, limiting the benefits of high temporal resolution of event cameras. More recent work, BlinkTrack~\cite{shen2025blinktrack}, improves robustness to occlusions and efficiency through better training data and filtering strategies.

Despite these advances, existing methods~\cite{han2025mate_track, wan2025event_track, burner2022evimo2, li2026moflow, luo2024efficient_flow, zhang2023frame_track, wang2024event_track, liu2025timetracker_track} rely on predefined criteria to determine the accumulation duration, which limits their ability to adapt to varying motion dynamics. In this work, we address this limitation by introducing a reinforcement learning-based approach that adaptively determines the temporal window for event accumulation, enabling robust tracking across diverse motion patterns. Furthermore, existing evaluation datasets~\cite{EC_mueggler2017event, EDS_hidalgo2022event} often exhibit relatively uniform motion patterns, limiting their ability to assess performance under diverse and dynamic conditions. To address this, we introduce a dynamic event-based tracking (DEFT) dataset, which captures realistic scenarios with diverse ever-changing motion distributions within each sequence.

\noindent
\textbf{Temporal Windowing of Event Streams.}
Processing event streams typically requires aggregating asynchronous events into temporal windows to construct dense event representations~\cite{maqueda2018event, zhu2019unsupervised, manderscheid2019speed, zubic2023chaos, fan2025eventpillars, xu2025mets} for downstream tasks such as object detection~\cite{wu2024leod, gehrig2023recurrent, kang2025unleashing, hamaguchi2023hierarchical, lu2026flexevent, ahmed2025efficient_od, peng2024scene_od, zhurethinking_flow, cho2025ev}, recognition~\cite{kim2021n, zhou2024exact_recog, zheng2024eventdance_recog, kim2022ev_recog, cho2023label_recog, plizzari2022e2_recog}, depth estimation~\cite{nam2022stereo, yan2025event_depth, ghosh2025event, zhu2025depth_depth, liang2025eventups_depth, cho2024temporal_depth, bartolomei2025depth_depth, yu2024eventps_depth, cho2023learning_depth} and scene reconstruction~\cite{liao2024ef_gs, yura2025eventsplat_gs, huang2025inceventgs_gs, yu2025evagaussians_gs, han2024event_gs}. A common strategy is to accumulate events over a fixed temporal interval or according to predefined criteria, enabling the use of conventional vision pipelines while simplifying the inherently continuous nature of event data.

However, the role of temporal windowing differs significantly across tasks. For tasks such as detection, predictions are made based on information within a given temporal window, and the choice of the window boundaries is relatively flexible. In contrast, feature tracking inherently estimates the motion of points over time, where each prediction corresponds to the displacement between consecutive states. 

As a result, the starting point of each temporal window is naturally defined by the previous estimation, making the choice of the accumulation interval fundamentally more constrained.
This property makes tracking particularly sensitive to the temporal extent of event accumulation. A short temporal window may fail to capture sufficient motion information, leading to unstable estimates, while a long temporal window can blur motion dynamics or degrade performance under non-linear motion. Therefore, under varying motion dynamics, fixed temporal windowing strategies are insufficient for robust event-based feature tracking.

\noindent
\textbf{Reinforcement Learning for Vision Tasks.} 
Reinforcement learning (RL) has been widely adopted in computer vision\cite{he2016deep_visualrl, song2018seednet_visualrl, caicedo2015active_visualrl, shang2023active} for problems where explicit supervision is unavailable or difficult to define. In tasks such as object detection~\cite{samiei2022object, novkovic2020object, pitcher2024reinforcement, mathe2016reinforcement_visualrl, ding2023learning_visualrl_od}, visual tracking~\cite{luo2018end, Messikommer24eccv, zhang2017deep_visualrl}, and active perception~\cite{cheng2018reinforcement, shang2023active, whitehead1990active, wang2025event, grimes2023learning, krueger2020activereinforcementlearningobserving}, the optimal action cannot be directly specified, and its quality is evaluated only through its impact on the final performance. As a result, models must learn appropriate actions without ground-truth labels for each decision, relying instead on feedback signals to optimize long-term objectives. RL provides a principled framework for addressing such settings by enabling policy learning from reward signals, and has been used to learn behaviors such as selecting informative regions~\cite{caicedo2015active, Messikommer24eccv}, controlling viewpoints~\cite{shang2023active}, and state control~\cite{Zhang2023InteractiveOP, Lee2024LearningTC, pinto2017asymmetric}, where the correctness of each action is inherently ambiguous.

This challenge naturally arises in event-based feature tracking. Given a continuous and asynchronous event stream, determining how much temporal information to accumulate for reliable inference does not have a well-defined ground-truth, and the quality of a given temporal window can only be assessed through the resulting tracking performance. However, existing methods rely on fixed heuristics for temporal aggregation, which limits their ability to adapt to varying motion dynamics.
In this work, we address this limitation by leveraging RL to adaptively determine the temporal window for event accumulation. This enables the model to learn when to perform inference without explicit supervision, leading to improved robustness under diverse motion conditions.

\section{Methodology}
\label{sec:method}

\begin{figure*}[t]
    \centering
    \includegraphics[width=.99\linewidth]{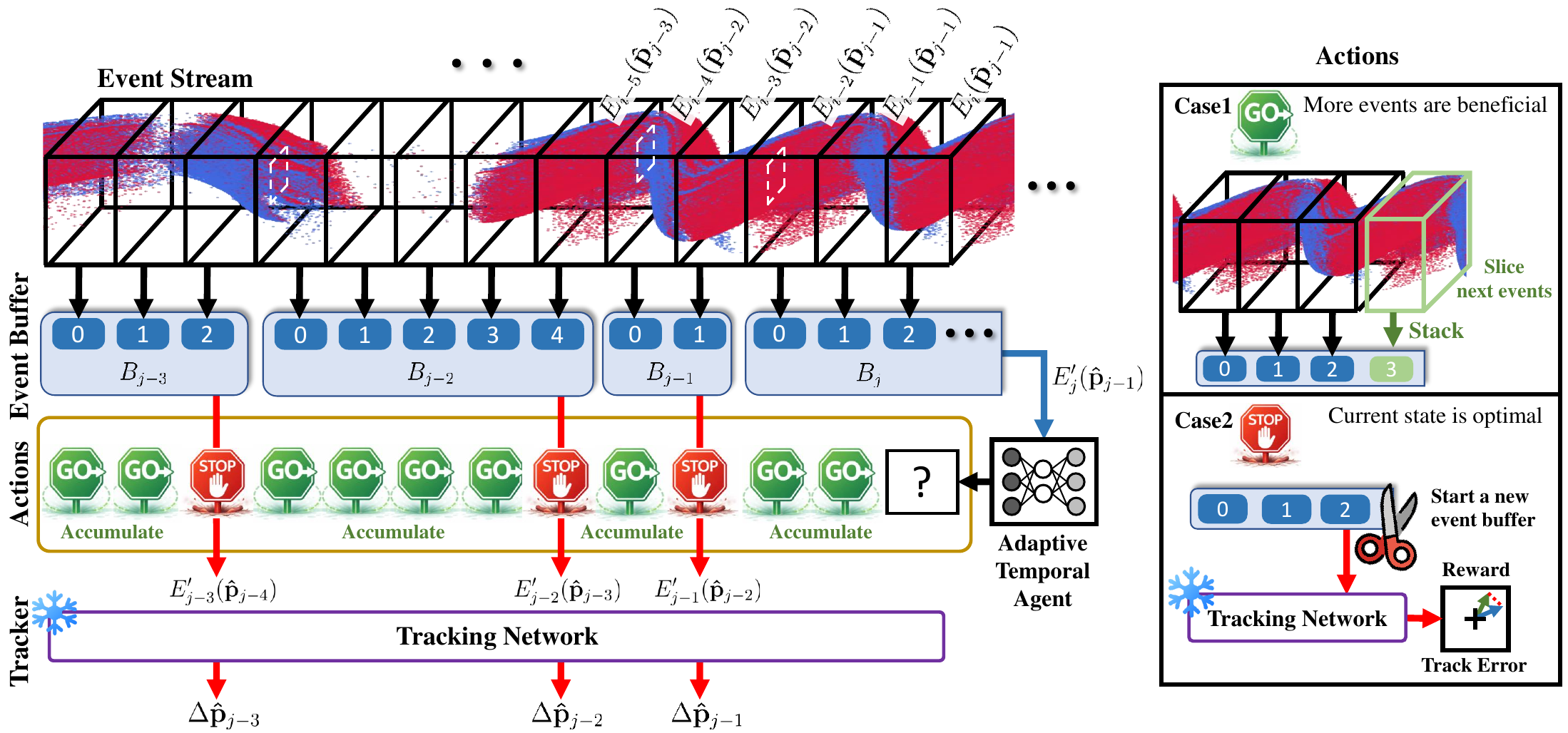}
    \vspace{-5pt}
    \caption{Overview of the proposed framework. Events are accumulated in a buffer, and the Adaptive Temporal Agent decides whether to trigger tracking inference. It performs inference when sufficient information is available, delays updates otherwise, and increases inference frequency under rapid motion for stable tracking.
    }
    \label{fig:overall_framework}
    \vspace{-10pt}
\end{figure*}

Despite extensive research on event-based feature tracking, maintaining stable and high tracking performance in real-world scenarios remains challenging due to motion-induced variability in triggered events and limited computational resources, particularly when balancing performance with computational efficiency.
To address this, we propose a RL–based approach in which a deep neural agent exploits the temporal dynamics of event data and adaptively accumulates events for robust tracking. The agent optimizes the tracker input online, effectively resolving the trade-off between inference time and tracking performance.
The remainder of this section is organized as follows. In Sec.~\ref{sec:subsec_feature}, we present the underlying tracking module with an adaptive temporal window, and in Sec.~\ref{sec:subsec_rl}, we describe the reinforcement learning framework for optimizing temporal aggregation of events.

\subsection{Event-based Feature Tracking with Adaptive Temporal Window} 
\label{sec:subsec_feature}

\noindent
\textbf{Event-based Feature Tracking.}
Conventional event-based feature tracking aims to estimate the motion of visual features within small patches. At current timestamp $t = t_i$, tracker receives event input $E_{i}(\mathbf{\hat{p}}_{i-1})$, generated within the temporal window $[t_{i-1} , t_{i}]$ of fixed length inside a square patch centered at $\mathbf{\hat{p}}_{i-1}= (x_{i-1}, y_{i-1})$. 
The model predicts the feature displacement $\Delta \mathbf{\hat{p}}_i = (\Delta \hat{x}_i, \Delta \hat{y}_i)$ and updates the patch center as $\mathbf{\hat{p}}_{i} = \mathbf{\hat{p}}_{i-1} + \Delta \mathbf{\hat{p}}_i$, which is then used to obtain the next input $E_{i+1}(\mathbf{\hat{p}}_{i})$. Because each prediction determines the input for subsequent steps, error between the predicted center and the ground-truth center accumulates over time.
To improve the robustness of event-based feature tracking, we propose a method that controls the temporal window at the input level of the event module. 
Conventional tracking approaches~\cite{shen2025blinktrack, messikommer2023data} use event data $E_i(\mathbf{\hat{p}}_{i-1})$ from a fixed temporal interval $T_i = [t_i - \Delta t, t_i]$ to predict motion at each timestamp $t_i$. 
In contrast, we decouple inference from fixed timestamps by introducing an inference index $j$, where each index corresponds to a tracker inference step. 
At each step $j$, the agent maintains a buffer $B_j$ that accumulates incoming event streams until sufficient motion information is observed. 
Instead of performing inference at every timestamp, the agent continuously aggregates events into the current buffer and evaluates motion cues from the buffered events. 
The agent triggers tracker inference only when the buffer contains sufficient information; once inference is triggered, the buffer is flushed and a new buffer $B_{j+1}$ is initialized for subsequent event accumulation.
At each inference step $j$, the tracker receives the event stream $E'_j(\mathbf{\hat{p}}_{j-1})$ within the optimized temporal window $[t_{j-1}, t_j]$ and outputs the displacement $\Delta \mathbf{\hat{p}}_j$ to update the patch center $\mathbf{\hat{p}}_j$. 
The temporal window is defined by the previous inference timestamp, $t_{j-1} = t_j - \Delta t \cdot l_j$, where $l_j$ denotes the number of event accumulations stored in $B_j$. 
To enable fine-grained control of the temporal window, we set $\Delta t$ to be half of that used in conventional trackers. 
By learning adaptive temporal windows, the agent dynamically adjusts the amount of event information according to scene motion, reducing overall inference time while improving tracking accuracy. 
This simple control mechanism enables asynchronous operation of the tracking module, consistent with the event camera's sensing paradigm based on intensity changes.

\noindent
\textbf{Sequential Decision Problem Formulation.}
For effective temporal window control, we introduce a reinforcement learning (RL) agent that decides whether to perform inference (\textit{stop}) using a pretrained event-based tracker or accumulate the next event stream (\textit{go}). The RL agent is designed to encode the tendency of triggered events and determine whether the accumulated event stream is sufficient for reliable motion estimation under varying event densities. 
We formulate temporal window control as a sequential decision-making problem, where the agent selects an action $a \in A$ at state $s \in S$. 
The state $s$ represents the events accumulated in the buffer $B_j$, and the action space $A$ consists of two discrete actions: \textit{go} and \textit{stop}. Each action is evaluated using a reward function $r(s,a)$, and future rewards are discounted by a factor $\gamma$.
Our agent is optimized using two networks, a policy network and a critic network. The policy network parameterizes the policy $\pi(a|s)$, which represents the probability distribution over actions given the current state. The critic network approximates the value function $V^{\pi}(s_i)$, which estimates the expected sum over the discounted rewards from the state $s_i$ over trajectories $\tau(\pi)$ induced by policy $\pi$. 
\begin{equation}
    V^{\pi}(s_i) = \mathbb{E}_{\tau(\pi)} {[\sum^{\infty}_{t=i} \gamma^t r(s_t, a_t)]}.
    \label{equ:policy}
\end{equation}

\subsection{Reinforcement Learning for Adaptive Temporal Agent}
\label{sec:subsec_rl}

\noindent
\textbf{State Design.}
Many prior works~\cite{yuan2022pre, seo2022reinforcement, liu2024visual, luo2024pre} employ feature embeddings extracted from pretrained networks as state representations for reinforcement learning agents.
A straightforward approach in our setting would be to leverage features from a pretrained tracking encoder as the state representation.
However, such intermediate representations introduce several limitations. 
First, features extracted from the tracking network are optimized for the tracking objective, which may not align with the decision-making process of the RL agent, thereby introducing irrelevant or misleading information.
Second, computing such features with complex encoders incurs additional computational overhead and latency, which is undesirable for high-rate event streams and can hinder timely decision making.

For effective decision-making under a limited computational budget, we design a lightweight architecture which enables fast decision-making and adopt the event representation~\cite{shen2025blinktrack} with the current accumulation information, providing the policy network with rich motion cues.
We encode the 2D event representation $E'_j(\mathbf{\hat{p}}_{j-1}) \in \mathbb{R}^{P \times P \times C}$ using 2D convolution layers, where $C$ denotes the number of event channels and $P$ denotes the patch size. The resulting feature vector is concatenated with the temporal window information and fed into a multi-layer perceptron (MLP) to produce the final action. This lightweight policy architecture reduces the overall inference time despite the additional computation introduced by the decision making.

\noindent
\textbf{Reward Design.}
The desirable behavior is accumulating event data until sufficient motion information is accumulated while avoiding excessive accumulation that may degrade tracking reliability. The agent must balance \textit{go} and \textit{stop} actions to trade off between frequent inference for accuracy and infrequent inference for efficiency. To achieve this, we design a simple and intuitive reward function:
\begin{equation}
     r = 
     \begin{cases}
     L_j \cdot \max{(-1,0.6 - e_i)} - \lambda & \text{if}~a_i = \textit{stop} \\
     0 & \text{if}~a_i =  \textit{go} \\
    \end{cases}, \quad
    e_i = {1 \over P} ||\mathbf{\hat{p}}_{i} - \mathbf{p}_{i}||_2
\label{equ:reward}
\end{equation}
where $e_i$ denotes the tracking error at current timestamp $t_i$ and $L_j$ denotes the information of event stacks in $B_j$. Incorporating this error into the reward encourages decisions that improve tracking performance. This formulation encourages the agent to continue accumulating events as long as the tracking module can reliably estimate motion. When the accumulated events are insufficient and noisy, the agent selects \textit{go} which increases $L_j$. Conversely, when the temporal window becomes excessively large, both $L_j$ and $e_i$ tend to increase, resulting in a large negative reward in the next \textit{stop}. 
However, this formulation may bias the policy toward excessive use of the \textit{stop} action. To prevent this behavior, we introduce an additional penalty term $\lambda$, which discourages overuse of \textit{stop} which can lead to unreliable estimation.

\noindent 
\textbf{Policy Optimization.}
We train the RL agent using the on-policy algorithm Proximal Policy Optimization (PPO)~\cite{schulman2017proximal}. PPO is a widely adopted policy gradient algorithm known for its training stability, which is achieved through conservative policy updates by clipping policy ratio. The policy function is primarily optimized by maximizing the following clipped surrogate objective:
 
\begin{equation}
\begin{aligned}
    L^{CLIP}(\theta) = \hat{\mathbb{E}}_t& [min(r_t(\theta)\hat{A_t}, clip(r_t(\theta), 1- \epsilon, 1+\epsilon) \hat{A}_t], \\
    &r_t(\theta) = {{\pi_{\theta} (a_t | s_t)} \over {\pi _{\theta_{old}}}(a_t | s_t)}.
    \label{equ:advantage}
\end{aligned}
\end{equation}
The advantage function $\hat{A_t}$ measures how beneficial action $a_t$ is at state $s_t$. It is estimated using Generalized Advantage Estimation (GAE)~\cite{schulman2015high} whose computation requires value predictions from the critic network. The advantage function evaluates action quality considering the difficulty of the state itself \cite{sutton1999policy}. When the advantage is positive, the gradient flows in the way that increases the probability $\pi_{\theta}(a=a_t | s=s_t)$. Meanwhile, a negative advantage reduces the probability of unreliable actions, leading to stable and efficient policy improvement.

\noindent
\textbf{Privileged Information for Critic.}
During the policy optimization, tracking failure may occur for two different reasons. 
In some cases, the failure is caused by an inappropriate temporal window due to the control failure of the RL agent. In other cases, however, the agent chooses a reasonable temporal window while the tracker still drifts to an incorrect feature due to factors such as occlusion or interference from nearby structures. 
In latter cases, the agent receives a large negative reward even though the failure is not caused by its decision.
Since the advantage estimate in Eq.~(\ref{equ:advantage}) is computed from the value function, such noisy rewards can introduce inconsistent value targets, which in turn destabilize the policy learning process. Although one could attempt to design a reward function considering these edge cases, such manual reward shaping may instead bias the agent toward unintended suboptimal behaviors.

For stable value function approximation, we provide the critic network with additional privileged information available \emph{only during training} to help distinguish the causes of tracking failure and take into account the inherent difficulty of the scene.
Specifically, we provide the critic with ground-truth motion information for the feature tracking task, along with the remaining timesteps in the sequence.
The ground-truth motion information includes the motion speed of the target feature, occlusion information and the tracking error, allowing the critic to infer the difficulty of the scene, such as rapid motion and occlusion. The remaining timesteps indicate how much time remains in the sequence, enabling the critic to better assess the long-term impact of the current action. 
Importantly, this privileged information is used only during training and is accessible solely to the critic, but not to the policy.
With this privileged information, well-fitted critic network effectively guides the policy network toward the desired behavior.

\begin{figure*}[t]
    \centering
    \includegraphics[width=.99\linewidth]{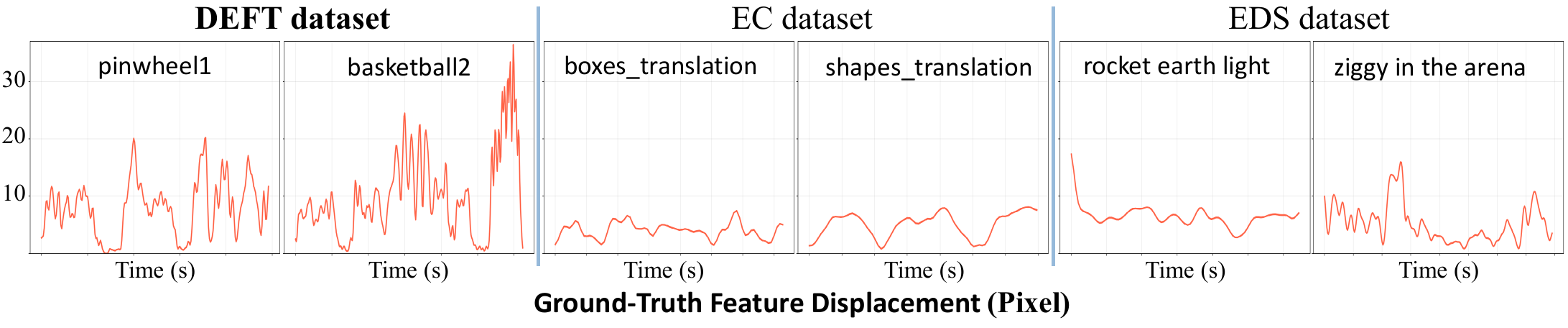}
    \vspace{-7pt}
    \caption{Average feature displacement over time for selected scenes, illustrating representative motion distributions. EDS is slightly more dynamic than EC, but both are largely uniform, whereas DEFT exhibits more diverse and abrupt motion patterns, highlighting the strengths of event cameras.
    }
    \label{fig:dataset_distribution}
    \vspace{-8pt}
\end{figure*}

\section{Dynamic Event-based Feature Tracking(DEFT) dataset}
Although real datasets~\cite{EDS_hidalgo2022event, EC_mueggler2017event} have been proposed for event-based feature tracking, the motion patterns in these datasets are often relatively simple and lack the diversity commonly observed in real-world scenarios. We observe that conventional tracking methods frequently fail when deployed in real environments with highly variable feature motion. To address this limitation, we introduce a new real-world event-based feature tracking dataset, the Dynamic Event-based Tracking dataset (DEFT), which also serves as a comprehensive benchmark for evaluation.
DEFT is captured using a DAVIS346C camera, which provides 346 $\times$ 260 asynchronous event streams. The dataset contains approximately 57 seconds of event stream data across eight scenarios which have diverse object motion. Ground-truth feature trajectories are manually annotated at 25 Hz.
As shown in Fig.~\ref{fig:dataset_distribution}, trajectories in DEFT exhibit significantly diverse motion patterns including abrupt stops, rapid accelerations, and frequent motion changes that produce highly dynamic event streams. In addition, the dataset contains heterogeneous motion from both foreground objects and background elements, making the tracking task substantially more challenging and establishing it as a strong benchmark for evaluating real-world event-based applications. More details on the DEFT dataset are provided in the supplementary material.

\section{Experiments}
\label{sec:experiment}

\subsection{Implementation Details} 
We train our agent using the on-policy algorithm Proximal Policy Optimization (PPO)~\cite{schulman2017proximal}, implemented in Stable-Baselines3~\cite{raffin2021stable}.
Training is performed on a single NVIDIA TITAN RTX 24GB GPU with 16 parallel environments and takes approximately 12 hours. 
For training, an event-stream sequence is clipped to 95 event frames.
The agent is trained for $5\times10^6$ iterations where one iteration denotes one state transition. We use the Adam optimizer~\cite{kingma2014adam} with a learning rate linearly decayed from 1e-4 to 1e-5, and set the discount factor $\gamma = 0.9$ so that our agent considers future rewards as well. The policy network and critic network consist of two CNN encoders with ReLU activations \cite{agarap2018deep}, which encode the event representation into a 2D feature vector, followed by three MLP layers that output the action and value function, respectively. We freeze the parameter of a pre-trained tracker during RL training. For fair comparison with conventional event-based tracking methods, we adopt event representation consistently with previous  works~\cite{messikommer2023data, shen2025blinktrack, wang2019event}. We set $\Delta t = 0.005s$ for high temporal resolution of the asynchronous tracking. In Eq.~(\ref{equ:reward}), we set $\lambda = 0.2$ for reward function and $L_j = 2\cdot l_j$.

\subsection{Experimental Setup}

\noindent
\textbf{Baselines.}
We evaluate our method in a plug-and-play manner on top of recent event-based feature trackings, including BlinkTrack~\cite{shen2025blinktrack} and its concurrent work Deep-EV-Tracker~\cite{messikommer2023data}.
By integrating our RL-based temporal window control module into these learnable trackers, we assess the performance gains achieved without modifying their underlying architectures.
We additionally compare against BlinkTrack augmented with a Kalman filter~\cite{kalman1960new}, which has been shown to improve tracking performance on the EC and EDS datasets.
Furthermore, we include learning-free baseline, HASTE~\cite{alzugaray2020haste}, which represent rule-based event feature tracking approaches.

\noindent
\textbf{Datasets.}
Prior event-based feature trackings typically adopt a cross-domain generalization setting, where training and evaluation datasets are separated.
In this work, we follow the same cross-domain setting and use the MultiTrack dataset~\cite{shen2025blinktrack} for training, as in prior work, to demonstrate the effectiveness of our method without introducing additional training data.
The tracker is adopted using pretrained models provided by the original authors, whose parameters are kept fixed during training, while the training dataset is used solely to learn the decision-making policy of the RL agent. We evaluate our method on three real-world event-based tracking datasets: DEFT, EC~\cite{EC_mueggler2017event}, and EDS~\cite{EDS_hidalgo2022event}.

\begin{table}[t]
\setlength{\tabcolsep}{1.6pt}
\centering
\caption{Performance comparison on the DEFT dataset. We report Feature Age (FA) and Expected Feature Age (EFA).
All methods use only event data. DT and BT denote Deep-EV-Tracker~\cite{messikommer2023data} and BlinkTrack~\cite{shen2025blinktrack}, respectively, and K.F denotes the Kalman Filter. Bold and underlined values indicate the best and second-best results, respectively. Values in parentheses indicate improvements over the base model.}
\vspace{-6pt}
\label{tab:fa_results}
\renewcommand{\arraystretch}{1.17}
\resizebox{\linewidth}{!}{
\begin{tabular}{l|cc|cc|cc|cc|cc|cc|cc|cc||cc}
\hline
\multirow{2}{*}{Method} & \multicolumn{2}{c|}{Desk} & \multicolumn{2}{c|}{Pinwheel1} & \multicolumn{2}{c|}{Pinwheel2} & \multicolumn{2}{c|}{Pinwheel3} & \multicolumn{2}{c|}{Basketball1} & \multicolumn{2}{c|}{Basketball2} & \multicolumn{2}{c|}{RobotArm1} & \multicolumn{2}{c||}{RobotArm2} & \multicolumn{2}{c}{Average}  \\
 & FA$\uparrow$ & EFA$\uparrow$ & FA$\uparrow$ & EFA$\uparrow$ & FA$\uparrow$ & EFA$\uparrow$ & FA$\uparrow$ & EFA$\uparrow$ & FA$\uparrow$ & EFA$\uparrow$ & FA$\uparrow$ & EFA$\uparrow$ & FA$\uparrow$ & EFA$\uparrow$ & FA$\uparrow$ & EFA$\uparrow$ & FA$\uparrow$ & EFA$\uparrow$ \\
\hline
HASTE~\cite{alzugaray2020haste} & 0.565 & 0.562 & 0.021 & 0.020 & 0.244 & 0.243 & 0.379 & 0.379 & 0.196 & 0.196 & 0.034  & 0.034 & 0.079 & 0.079 & \underline{0.435} & \underline{0.435} & 0.244 & 0.243 \\
\hline
\multirow{2}{*}{DT~\cite{messikommer2023data}} & \multirow{2}{*}{0.172} & \multirow{2}{*}{0.170} & \multirow{2}{*}{0.016} & \multirow{2}{*}{0.016} & \multirow{2}{*}{0.131} & \multirow{2}{*}{0.131} & \multirow{2}{*}{0.135} & \multirow{2}{*}{0.134} & \multirow{2}{*}{0.052} & \multirow{2}{*}{0.052} & \multirow{2}{*}{0.024} & \multirow{2}{*}{0.024} & \multirow{2}{*}{0.061} & \multirow{2}{*}{0.060} & \multirow{2}{*}{0.092} & \multirow{2}{*}{0.091} & \multirow{2}{*}{\thead{0.086 \\ (-)}} & \multirow{2}{*}{\thead{0.085 \\ (-)}}  \\
& & & & & & & & & & & & & & && &  & 
\\
\gr
& & & & & & & & & & & & & & && &  &  \\
\gr
\multirow{-2}{*}{DT+\textbf{Ours}} & \multirow{-2}{*}{\underline{0.590}} & \multirow{-2}{*}{\underline{0.584}} & \multirow{-2}{*}{\underline{0.298}} & \multirow{-2}{*}{0.279} & \multirow{-2}{*}{0.391} & \multirow{-2}{*}{\underline{0.387}} & \multirow{-2}{*}{\underline{0.398}} & \multirow{-2}{*}{\underline{0.398}} & \multirow{-2}{*}{\underline{0.697}} & \multirow{-2}{*}{\underline{0.696}} & \multirow{-2}{*}{0.370} & \multirow{-2}{*}{0.370} & \multirow{-2}{*}{\underline{0.572}} & \multirow{-2}{*}{\underline{0.572}} & \multirow{-2}{*}{0.399} & \multirow{-2}{*}{0.396} &  \multirow{-2}{*}{\thead{\underline{0.465} \\ (+0.379)}}  & \multirow{-2}{*}{\thead{\underline{0.460} \\ (+0.375)}} \\

\hline
\multirow{2}{*}{BT~\cite{shen2025blinktrack}} & \multirow{2}{*}{0.157} & \multirow{2}{*}{0.157} & \multirow{2}{*}{0.293} & \multirow{2}{*}{\underline{0.293}}  & \multirow{2}{*}{0.398} & \multirow{2}{*}{0.382} & \multirow{2}{*}{0.188} & \multirow{2}{*}{0.186} & \multirow{2}{*}{0.644} & \multirow{2}{*}{0.639} & \multirow{2}{*}{0.488} & \multirow{2}{*}{0.478} & \multirow{2}{*}{0.557} & \multirow{2}{*}{0.541} & \multirow{2}{*}{0.300} & \multirow{2}{*}{0.300} & \multirow{2}{*}{\thead{0.378 \\ (-)}} & \multirow{2}{*}{\thead{0.372 \\ (-)}} \\
& & & & & & & & & & & & & & && & &
\\
\multirow{2}{*}{BT+K.F~\cite{shen2025blinktrack}}& \multirow{2}{*}{0.157} & \multirow{2}{*}{0.157} & \multirow{2}{*}{0.245} & \multirow{2}{*}{0.245} & \multirow{2}{*}{\underline{0.400}} & \multirow{2}{*}{0.384} & \multirow{2}{*}{0.182} & \multirow{2}{*}{0.181} & \multirow{2}{*}{0.669} & \multirow{2}{*}{0.664} & \multirow{2}{*}{\underline{0.510}} & \multirow{2}{*}{\underline{0.500}} & \multirow{2}{*}{0.558} & \multirow{2}{*}{0.543} & \multirow{2}{*}{0.295} & \multirow{2}{*}{0.294} & \multirow{2}{*}{\thead{0.377 \\ ($-$0.001)}} & \multirow{2}{*}{\thead{0.371 \\ ($-$0.001)}}  \\
& & & & & & & & & & & & & & & &&  &  \\
\gr
& & & & & & & & & & & & & & && &   &  \\
\gr
\multirow{-2}{*}{BT+\textbf{Ours}} & \multirow{-2}{*}{\textbf{0.836}} & \multirow{-2}{*}{\textbf{0.835}} & \multirow{-2}{*}{\textbf{0.405}} & \multirow{-2}{*}{\textbf{0.405}} & \multirow{-2}{*}{\textbf{0.639}} & \multirow{-2}{*}{\textbf{0.635}} & \multirow{-2}{*}{\textbf{0.460}} & \multirow{-2}{*}{\textbf{0.459}} & \multirow{-2}{*}{\textbf{0.809}} & \multirow{-2}{*}{\textbf{0.808}} & \multirow{-2}{*}{\textbf{0.638}} & \multirow{-2}{*}{\textbf{0.634}} & \multirow{-2}{*}{\textbf{0.739}} & \multirow{-2}{*}{\textbf{0.738}} & \multirow{-2}{*}{\textbf{0.608}} & \multirow{-2}{*}{\textbf{0.608}} & \multirow{-2}{*}{\thead{\textbf{0.642} \\ (+0.264)}} & \multirow{-2}{*}{\thead{\textbf{0.640} \\ (+0.268)}} \\

\hline
\end{tabular}
}
\label{tab:main_det}
\vspace{-14pt}
\end{table}

\begin{figure*}[t]
    \centering
    \includegraphics[width=.99\linewidth]{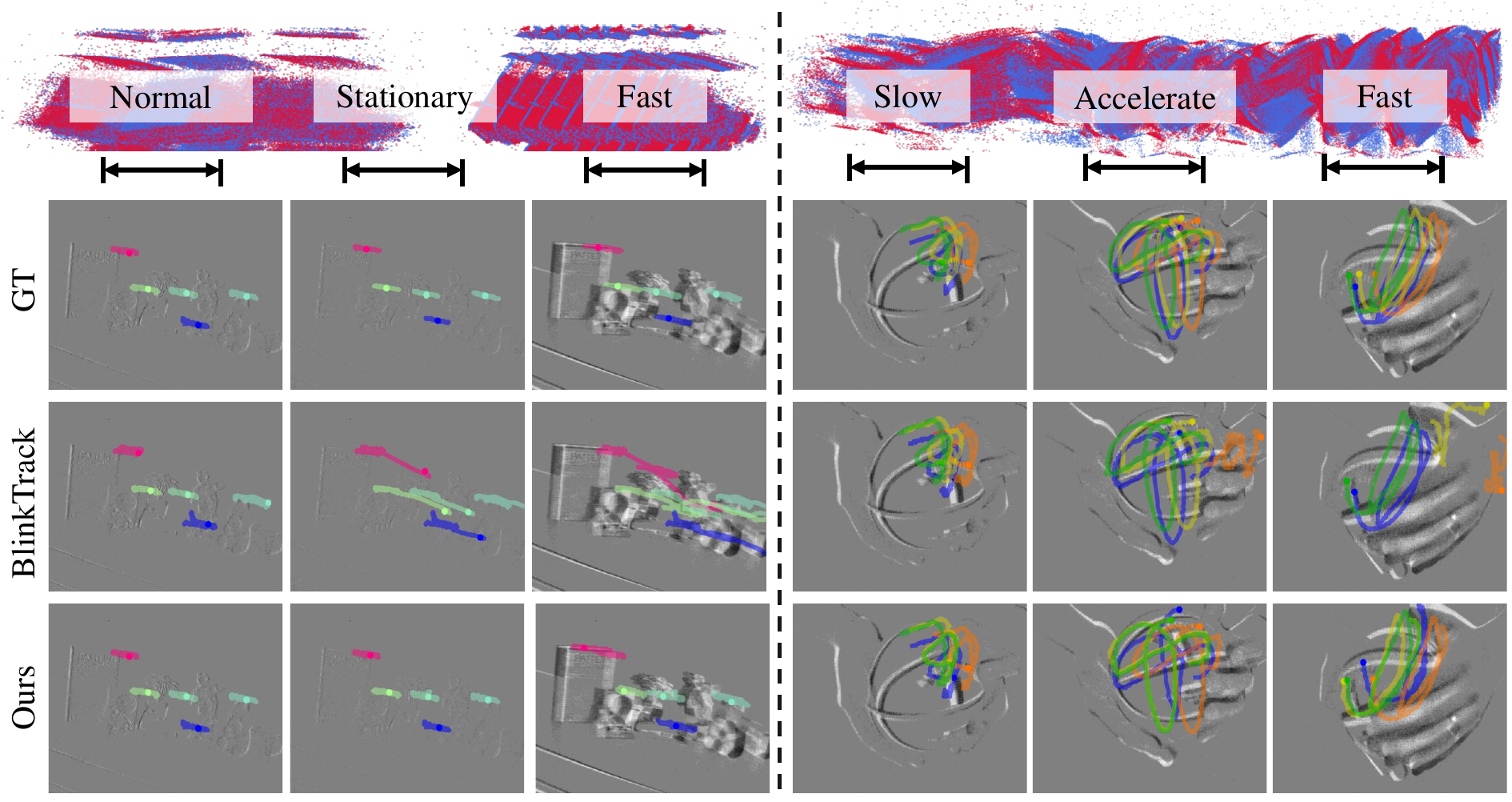}
    \vspace{-5pt}
    \caption{Qualitative comparison on the DEFT dataset. For clarity, feature tracking results are visualized over the corresponding interval, and only the most recent events within the current duration are shown.}
    \label{fig:qual_results}
    \vspace{-5pt}
\end{figure*}

\noindent
\textbf{Metric.} We follow the evaluation protocol of previous event-based feature trackings~\cite{shen2025blinktrack, messikommer2023data}. Since the EC, EDS, and DEFT datasets provide ground-truth annotations at a lower temporal resolution than the event-based predictions, we interpolate the predicted trajectories to the timestamps of the ground-truth annotations and compute the tracking error in pixel space.
We report two standard metrics: Feature Age (FA) and Expected Feature Age (EFA). Feature Age measures the proportion of time during which the tracking error remains below a predefined threshold~\cite{messikommer2023data}. Expected Feature Age is defined as the product of feature age and the ratio of stable tracks whose distance remains below the threshold across all predictions.

\subsection{Experimental Results}

\noindent
\textbf{Results on DEFT.} Table~\ref{tab:main_det} presents tracking performance under varying motion distributions, covering a wide range of motion dynamics.
Both Deep-EV-Tracker (DT)~\cite{messikommer2023data} and BlinkTrack (BT)~\cite{shen2025blinktrack} struggle under challenging conditions, and in certain scenes, their performance degrades significantly, sometimes failing to produce reliable tracking results.
When motion changes abruptly within a sequence, the tracker often loses reliable features, resulting in low average performance with FA of 0.086 and EFA of 0.085. Incorporating the proposed adaptive temporal agent improves robustness, allowing the model to handle challenging sequences, increasing FA and EFA by 0.379 and 0.375 on average.
BT is relatively more robust than DT and performs reasonably well across several sequences. However, it still struggles under highly dynamic motion and shows limited overall performance. Even when combined with a Kalman Filter \cite{shen2025blinktrack, kalman1960new}, the model cannot effectively handle such motion patterns. In contrast, integrating the proposed method with BT consistently improves tracking in all sequences, increasing FA and EFA by 0.264 and 0.268.
These results indicate that the proposed RL-based temporal controller 
adaptively shapes the raw event stream into informative temporal segments based on the distribution of triggered events, resulting in more reliable and accurate feature tracking.
We visualize the qualitative comparisons in Fig. \ref{fig:qual_results}. 
BT struggles in real-world scenarios with highly variable motion. In particular, it often fails when a moving object suddenly stops or abruptly accelerates, resulting in sparse or ambiguous event inputs. In contrast, our method adaptively adjusts the temporal window, enabling robust tracking under abrupt motion changes as well as both sparse and highly dense event conditions.

\noindent
\textbf{Results on EC and EDS.} 
Table~\ref{tab:eds_ec} compares our method with existing approaches on the EC and EDS, which mainly contain relatively monotonic motion patterns. 
On these benchmarks, ours improves the average performance of both trackers.
With the proposed method, DT achieves the highest score among all methods on the EC dataset, demonstrating a substantial performance gain. This result suggests that the proposed adaptive aggregation improves performance not only in challenging scenarios but also in relatively monotone scenarios.

\begin{table}[t]
\setlength{\tabcolsep}{11.2pt}
\centering
\caption{Performance on the conventional EDS and EC benchmarks. Higher is better.}
\vspace{-6pt}
\label{tab:fa_results}
\renewcommand{\arraystretch}{1.1}
\resizebox{0.9\linewidth}{!}{
\begin{tabular}{l|cc|cc||cc}
\hline
\multirow{2}{*}{Method} & \multicolumn{2}{c|}{EDS} & \multicolumn{2}{c||}{EC} & \multicolumn{2}{c}{Average} \\
 & FA$\uparrow$ & EFA$\uparrow$ & FA$\uparrow$ & EFA$\uparrow$ & FA$\uparrow$ & EFA$\uparrow$ \\
\hline
ICP~\cite{kueng2016low_icp} & 0.060 & 0.040 & 0.256 & 0.245 & 0.158 & 0.143\\
EM-ICP~\cite{zhu2017event_emicp} & 0.161 & 0.120 & 0.337 & 0.334 & 0.249	& 0.227\\
HASTE~\cite{alzugaray2020haste} & 0.096 & 0.063 & 0.442 & 0.427 & 0.269 & 0.245\\
\hline
DT~\cite{messikommer2023data} & 0.549 & 0.451 & 0.795 & 0.787 & 0.672 (-) & 0.619 (-)\\
\gr
\textbf{DT}~\cite{messikommer2023data}\textbf{+Ours} 
& \underline{0.622} & \underline{0.513} & \textbf{0.864} & \textbf{0.858} & \textbf{0.743} (+0.071) & \textbf{0.686} (+0.067)
\\
\hline
BT~\cite{shen2025blinktrack} & 0.568 & 0.474 & 0.833 & 0.819 & 0.701 (-) & 0.647 (-) \\
BT~\cite{shen2025blinktrack}$+$K.F & 0.569 & 0.475 & \underline{0.835} & \underline{0.820} & 0.702 (+0.001) & 0.648 (+0.001) \\
\gr
\textbf{BT}~\cite{shen2025blinktrack}\textbf{+Ours} & \textbf{0.649} & \textbf{0.534} & 0.819 & 0.811 & \underline{0.734} (+0.033) & \underline{0.672} (+0.025) \\
\hline
\end{tabular}
}
\label{tab:eds_ec}
\vspace{-12pt}
\end{table}

\subsection{Ablation Study and Analysis}

We evaluate the effectiveness of the proposed method through experiments using BlinkTrack on the DEFT dataset, along with ablation studies and analyses.

\noindent
\textbf{State Design Analysis.} 
Table~\ref{tab:state_rep} presents an analysis of the state representations used as input to the policy network.
We consider three types of state representations: event representation, features extracted from the tracker encoder and the output of the correlation layer in the tracker.
As shown in the results, using event representation yields the best performance.
We attribute this to the fact that features extracted by the tracking encoder are optimized for the tracking objective, which may not be well aligned with the decision-making process of the RL agent. In contrast, event representation preserves more direct and informative cues for policy learning. As a result, the policy network can effectively learn the decision-making process without relying on additional encoders, enabling a lightweight and efficient framework.

\noindent
\textbf{Action Penalty Analysis.} 
We conduct experiments on the action penalty coefficient $\lambda$ in Eq.~(\ref{equ:reward}), which discourages excessive use of the \emph{stop} action.
As shown in Table ~\ref{tab:action_penalty}, without the penalty term, the agent tends to overuse the \emph{stop} action, leading to suboptimal tracking performance.
Introducing the penalty encourages a better balance between \emph{go} and \emph{stop}, allowing the tracker to make predictions only after sufficient event information has been accumulated.
In particular, setting $\lambda = 0.2$ achieves a favorable trade-off between \emph{go} and \emph{stop} actions.

\noindent
\textbf{Comparison between RL Agent and Diverse Time-Interval Training.}
One possible way to improve robustness to dynamic and diverse motion distributions is to expose the model to a wide range of temporal intervals during training. To examine this, we extend the original training setting by dynamically sampling temporal intervals within each sequence, ranging from $\times$0.5 to $\times$2.5 the base interval. We then compare this strategy with the proposed RL agent.
As shown in Table ~\ref{tab:aug}, such time-interval augmentation does not effectively resolve the observed failure cases. In fact, training with a broad distribution of temporal intervals generally degrades overall performance. This suggests that simply increasing temporal diversity during training is insufficient, highlighting the advantage of the proposed adaptive temporal agent.

\begin{table*}[t]
\centering
\setlength{\tabcolsep}{4pt}
\renewcommand{\arraystretch}{1.1}
\begin{minipage}[c]{0.31\textwidth}
\centering
\captionsetup{type=table}
\caption{Analysis of the effect of state representation on performance.}
\label{tab:state_rep}
\setlength{\tabcolsep}{3.7pt}
\vspace{-5pt}
\resizebox{\linewidth}{!}{
\begin{tabular}{c|cc}
\hline
\multirow{2}{*}{\thead{State \\ Representation}} & \multirow{2}{*}{FA$\uparrow$} & \multirow{2}{*}{EFA$\uparrow$} \\
 & & \\
\hline
\thead{Event Representation} & \textbf{0.642} & \textbf{0.640} \\
\thead{Encoder Features} & 0.308 & 0.306\\
\thead{Correlation} & 0.171 & 0.169 \\
\hline
\end{tabular}}
\end{minipage}
\hfill
\begin{minipage}[c]{0.31\textwidth}
\setlength{\tabcolsep}{11pt}
\centering
\captionsetup{type=table}
\caption{Analysis of action penalty, $\lambda$ in Eq.~(\ref{equ:reward}).}
\label{tab:action_penalty}
\vspace{-3pt}
\resizebox{\linewidth}{!}{
\begin{tabular}{c|cc}
\hline
\multirow{2}{*}{\thead{Action \\ Penalty}} & \multirow{2}{*}{FA$\uparrow$} & \multirow{2}{*}{EFA$\uparrow$} \\
 & & \\
\hline
0 & 0.307 & 0.295 \\
0.1 & 0.600 & 0.595 \\
0.2 & \textbf{0.642} & \textbf{0.640} \\
0.3 & 0.614 & 0.611 \\
\hline
\end{tabular}}
\end{minipage}
\hfill
\begin{minipage}[c]{0.31\textwidth}
\setlength{\tabcolsep}{4.9pt}
\centering
\captionsetup{type=table}
\caption{Comparison of our method with data augmentation methods.}
\vspace{-3pt}
\renewcommand{\arraystretch}{1.2}
\label{tab:aug}
\resizebox{\linewidth}{!}{
\begin{tabular}{c|cc}
\hline
\multirow{2}{*}{Setting} & \multirow{2}{*}{FA$\uparrow$} & \multirow{2}{*}{EFA$\uparrow$} \\
 & & \\
\hline
BlinkTrack~\cite{shen2025blinktrack} & 0.378 & 0.372 \\
BlinkTrack w/ Aug. & 0.191 & 0.189 \\
BlinkTrack + \textbf{Ours} & \textbf{0.642} & \textbf{0.640} \\
\hline
\end{tabular}}
\end{minipage}
\vspace{-5pt}
\end{table*}

\begin{table*}[t]
\centering
\begin{minipage}[t]{0.47\textwidth}
\centering
\captionsetup{type=table}
\renewcommand{\arraystretch}{1.2}
\setlength{\tabcolsep}{6.5pt}
\caption{Ablation study on privileged information for the critic network.}
\vspace{-12pt}
\label{tab:privileged}
\resizebox{0.98\linewidth}{!}{
\begin{tabular}{c|c|cc}
\hline
\multicolumn{2}{c|}{Privileged  Information} &  &  \\
\cline{1-2}
\multirow{2}{*}{\thead{GT Motion\\  Information}} & \multirow{2}{*}{\thead{ Remaining \\ Timestep}} & \multirow{2}{*}{FA$\uparrow$} & \multirow{2}{*}{EFA$\uparrow$} \\
& & &  \\
\hline
 & &0.456 & 0.455 \\
& \cmark & 0.491 & 0.490\\
\cmark& & 0.572 & 0.567 \\
\cmark & \cmark & \textbf{0.642} & \textbf{0.640} \\
\hline
\end{tabular}}
\end{minipage}
\hfill
\begin{minipage}[t]{0.46\textwidth}
\centering
\captionsetup{type=figure}
\includegraphics[width=.96\linewidth]{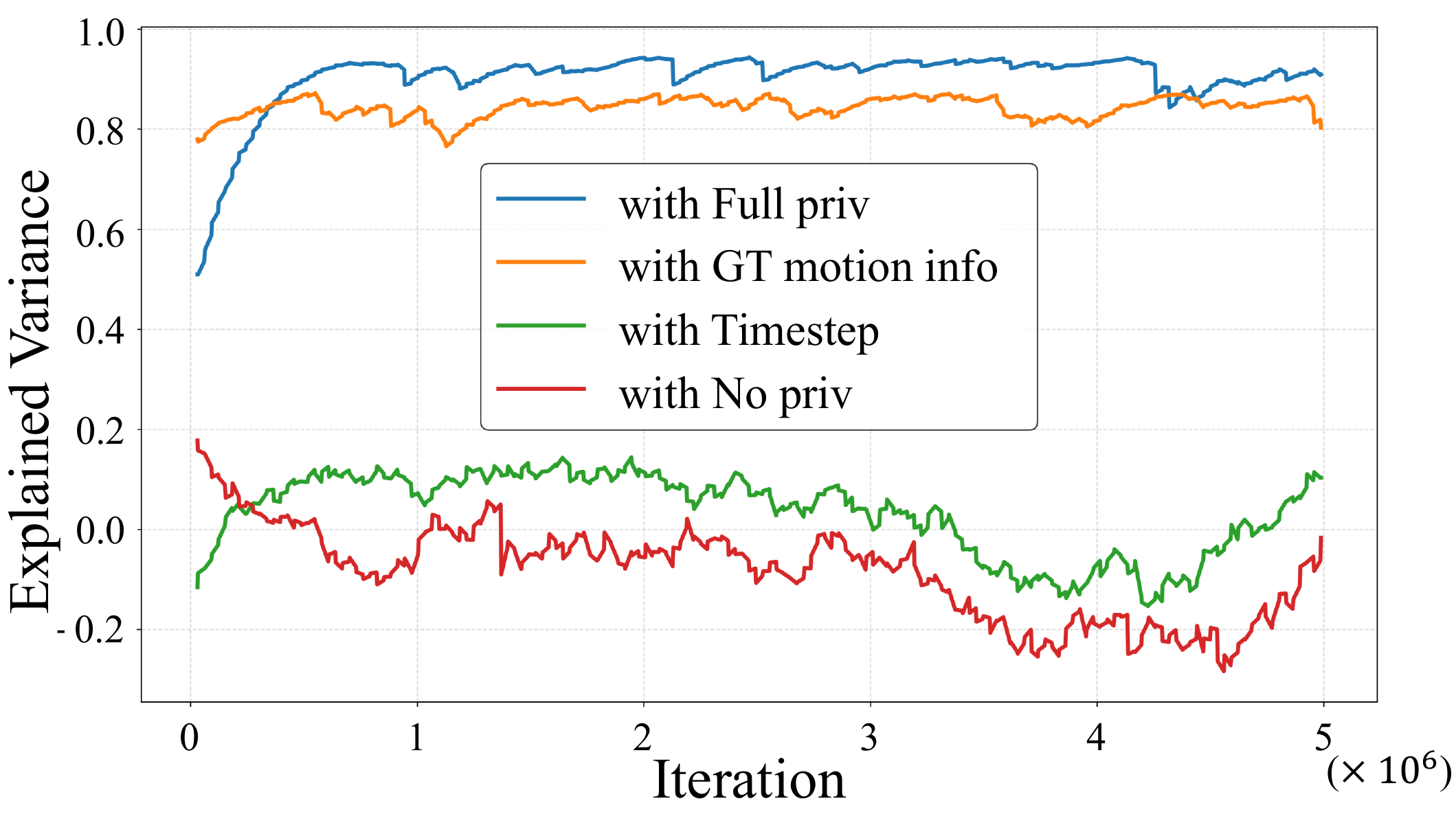}
\vspace{-2pt}
\caption{Training curves of the explained variance with and without privileged information.}
\label{fig:value_graph}
\end{minipage}
\vspace{-25pt}
\end{table*}

\noindent
\textbf{Privileged Information for the Critic Network.} We conduct an ablation study on the privileged inputs provided to the critic network. Specifically, we progressively remove each type of privileged information from the critic input and report the resulting performance degradation in Table ~\ref{tab:privileged}. The results show that each privileged component contributes to the overall performance gain. As shown in Fig.~\ref{fig:value_graph}, privileged information helps the critic network better approximate the expected return, as indicated by the higher explained variance.
As discussed in prior work~\cite{vapnik2009new, salter2021attention}, this demonstrates that a more accurate value function provides stronger guidance for the policy network to select actions that maximize the expected return.

\noindent
\textbf{Runtime Analysis.} 
Our RL agent requires approximately 0.4 ms (2,500 FPS) per preprocessed event patch to decide \textit{go} or \textit{stop} on a single NVIDIA TITAN RTX. Under the same hardware and experimental settings, the tracker module \cite{shen2025blinktrack} requires 5.3 ms (188 FPS). This shows that the proposed policy network introduces only negligible inference overhead. Owing to its lightweight design, the policy network effectively reduces unnecessary computations in the tracking module.

\noindent
\textbf{Adaptive Temporal Policy vs. Search-based Static Strategies.}
In general, prior studies~\cite{shen2025blinktrack, messikommer2023data} adopt a fixed temporal interval and perform inference at regular time steps for simplicity and computational efficiency. As shown in Table ~\ref{tab:rule_base}, we extend these methods to various temporal intervals and also consider number-based event stacking~\cite{wang2019event}.
For time-based stacking, performance varies significantly with the chosen interval, indicating that tracking accuracy is highly sensitive to this parameter and that selecting an optimal value is non-trivial. For number-based stacking, inference is triggered once a fixed number of events is accumulated. 
While this can improve robustness, it leads to excessive updates under high event density, causing a substantial increase in runtime and making it impractical for real-time applications. In particular, processing events corresponding to 1 second may take over 3–7 seconds, making it difficult to operate as an online feature tracking system. Such delays continuously accumulate over time, further degrading practical usability.
In contrast to the base setting ($\dagger$) with a fixed 0.01-second interval, our method adaptively shortens the inference interval under rapid motion, enabling successful tracking in scenarios where the base setting often fails. Although this increases the runtime due to more frequent updates, our method remains within the 1-second event segment duration, making it practical for online operation. Furthermore, when the event density is low, the inference frequency is reduced, resulting in even lower runtime than the base setting.

\section{Conclusion}
\label{sec:conclusion}

\begin{table}[t]
\setlength{\tabcolsep}{6.2pt}
\centering
\caption{Comparison of the trade-off between performance and runtime across various rule-based slicing strategies and our method on the DEFT dataset. Runtime is measured as the inference time required to process a single event-patch subsequence of 1 second. For time-based slicing, inference is performed at fixed temporal intervals, resulting in identical runtime regardless of the input event distribution. For number-based event slicing, runtime varies significantly depending on the number of events in each segment. To account for this variability, we report runtimes measured on segments whose event counts fall within the top 10\% and top 50\% of the entire dataset. $\dagger$ denotes the base setting of BlinkTrack~\cite{shen2025blinktrack}.}
\vspace{-6pt}
\label{tab:fa_results}
\renewcommand{\arraystretch}{1.1}
\resizebox{0.99\linewidth}{!}{
\begin{tabular}{c|ccccc|ccccc||c}
\hline
\multirow{2}{*}{Metrics} & \multicolumn{5}{c|}{Number (Count) } & \multicolumn{5}{c||}{Time (Second)} & \multicolumn{1}{c}{\textbf{Ours}} \\
\cline{2-12}
 & 100 & 250 & 500 & 1000 &  2500 & 0.001 & 0.005 & \gc0.01$\dagger$ &  0.05 & 0.1 & \gcc Adaptive \\
\hline
FA$\uparrow$ & 0.542 & 0.580 & 0.585 & 0.461 & 0.207 & 0.100 & 0.121 & \gc 0.378 & 0.384 & 0.209 & \gcc 0.642  \\  
EFA$\uparrow$ & 0.540 & 0.577 & 0.582 & 0.458 &  0.205 & 0.043 & 0.108 & \gc 0.372  & 0.383 & 0.209 & \gcc 0.640 \\
\hline
Runtime (s) $\downarrow$ & \multirow{2}{*}{18.111} & \multirow{2}{*}{7.084} & \multirow{2}{*}{3.611} & \multirow{2}{*}{1.780}  & \multirow{2}{*}{0.360}  &  \multirow{4}{*}{5.436}  & \multirow{4}{*}{1.046}   &  \gc &  \multirow{4}{*}{0.107}  & \multirow{4}{*}{0.058}   & \gcc  \\ 
$[$Top 10\% Events$]$ & & & & & & & & \multirow{-2}{*}{\gc}  & & & \multirow{-2}{*}{\gcc 0.812} \\
Runtime (s) $\downarrow$ &   \multirow{2}{*}{3.080} & \multirow{2}{*}{1.196} & \multirow{2}{*}{0.631} & \multirow{2}{*}{0.318} & \multirow{2}{*}{0.150} &   &    &  \gc  &    & \   & \gcc  \\
$[$Top 50\% Events$]$ & & & & & & & & \multirow{-4}{*}{\gc 0.528} & & & \multirow{-2}{*}{\gcc 0.516}\\
\hline
\end{tabular}
}
\label{tab:rule_base}
\vspace{-12pt}
\end{table}

In this paper, we present a RL-based event feature tracking framework that adapts to the varying amount of event information encountered in real-world scenarios. With the adaptive temporal agent that dynamically controls the temporal window, the tracking module achieves both improved accuracy and reduced inference time across the real event dataset. Our learning-based adaptive temporal window consistently outperforms the hand-crafted rule-based alternatives in both practicality and efficiency. In addition, we present DEFT, our novel event-based feature tracking dataset designed to capture variable motion patterns in real-world scenarios. 
We hope that our RL-based temporal control framework encourages broader research of event-based robust perception in real-world applications.

\section{Acknowledgments.}

This work was supported by the Institute of Information \& communications Technology Planning \& Evaluation (IITP) grant funded by the Korea government(MSIT) (No. RS-2024-00457882, AI Research Hub Project), by the InnoCORE program of the Ministry of Science and ICT(N10250156), and by the National Research Foundation of Korea(NRF) grant funded by the Korea government(MSIT) (RS-2026-25473963).

%
%
\bibliographystyle{splncs04}
\bibliography{main}
\end{document}



\title{GoStop: Reinforcement Learning for \\ Adaptive Temporal Aggregation in \\ Event-Based Feature Tracking \\
\centering{\textit{---Supplementary Material---}}}

\titlerunning{GoStop}

\author{Youngho Kim$^{*}$\orcidlink{0009-0005-4024-5031} \and
Hoonhee Cho$^{*}$\orcidlink{0000-0003-0896-6793} \and
Jae-Young Kang\orcidlink{0009-0002-9537-3813} \and
Kuk-Jin Yoon\orcidlink{0000-0002-1634-2756}}
\authorrunning{Y.~Kim et al.}

\institute{KAIST 
\\
\email{\{kmax2001, gnsgnsgml, kjyoon\}@kaist.ac.kr}\\
}
\maketitle

In this supplementary material, we offer more details of our work, \textbf{GoStop}. Specifically, we provide
\begin{itemize}
\item Motivation for RL in event-based feature tracking in Section~\ref{sec:motivation};
\item Video demo in Section~\ref{sec:video_demo};
\item Details about DEFT dataset in Section~\ref{sec:det_data};
\item Experiments on other event representations in Section~\ref{sec:robust_event_repre};
\item More qualitative results in Section~\ref{sec:more_qual};
\item Analysis of the RL Module behavior according to the motion distribution in Section~\ref{sec:gostop};
\end{itemize}
\vspace{10pt}

\section{Motivation for RL in Event-based Feature Tracking}
\label{sec:motivation}

In the introduction of the main paper, we motivated the use of reinforcement learning for event-based feature tracking from the perspective of a decision-making problem. In this supplementary material, we further elaborate on the motivation and significance of introducing RL.

\noindent
\textbf{Suitability for Online Feature Tracking.}
As discussed in prior work~\cite{shen2025blinktrack, messikommer2023data}, event-based feature tracking aims to fully exploit the low-latency and high-temporal-resolution properties of event cameras, with the goal of operating in an \underline{online manner}. In other words, the tracker is expected to respond immediately as new events arrive. Although several previous methods~\cite{peng2023better, cao2024spiking} have proposed adaptive event slicing strategies, such approaches are generally designed for offline settings and therefore cannot be directly applied to online feature tracking problem. In other words, such methods operate in an offline manner, deciding how to partition data only after it has already been collected in the model.

More specifically, feature tracking problem requires the model to determine, at each moment after the most recent tracking inference, whether the accumulated events contain sufficient information and whether performing inference at the current time is optimal. In contrast, existing adaptive slicing methods focus on how to partition events that have already been accumulated, which is inherently an offline formulation. For this reason, such approaches are not directly applicable to online event-based feature tracking problem setting.

\noindent
\textbf{Balances Performance and Runtime.}
In feature tracking, a shorter inference interval typically leads to higher runtime, since achieving the same overall result requires more frequent tracker inference, which may in turn cause additional delay.
Thus, optimizing only for tracking performance can improve accuracy while significantly increasing computational cost. Our RL framework addresses this trade-off by incorporating a penalty term into the reward, enabling the model to balance performance and runtime automatically. This makes RL particularly suitable for decision-making problems where explicit ground-truth supervision is not available, while alternative solutions are often less straightforward.

\section{Video Demo}
\label{sec:video_demo}
To better demonstrate the feature tracking results, we also provide a supplementary video demo, through which the robustness of the proposed method can be clearly observed.

\section{Details about DEFT Dataset}
\label{sec:det_data}

\begin{figure*}[p]
    \centering
    \includegraphics[width=0.99\linewidth]{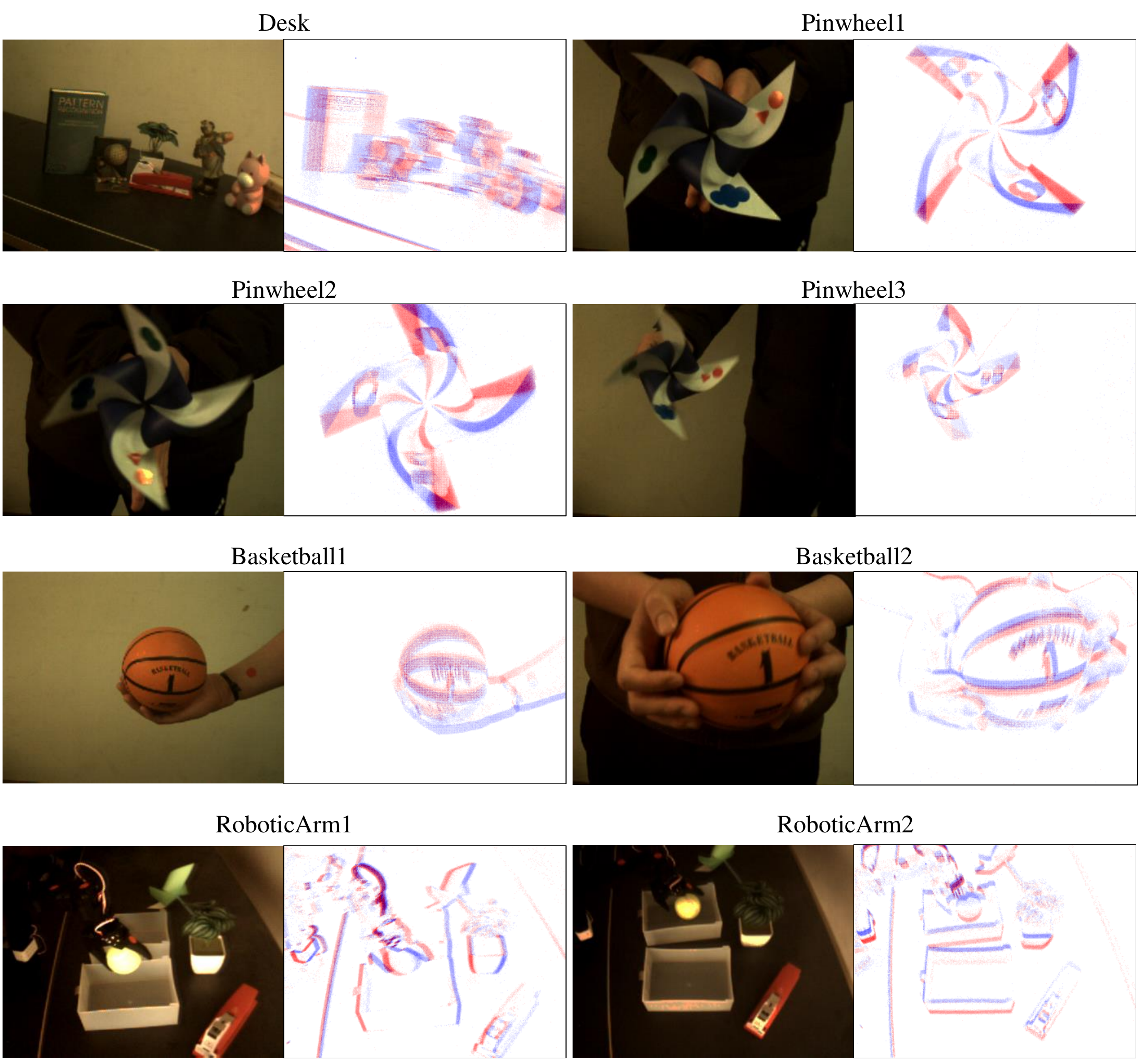}
    \vspace{-5pt}
    \caption{Visualization of the samples from DEFT dataset.}
    \label{fig:det_sample}
    \vspace{-10pt}
\end{figure*}

\begin{figure*}[p]
    \centering
    \includegraphics[width=0.99\linewidth]{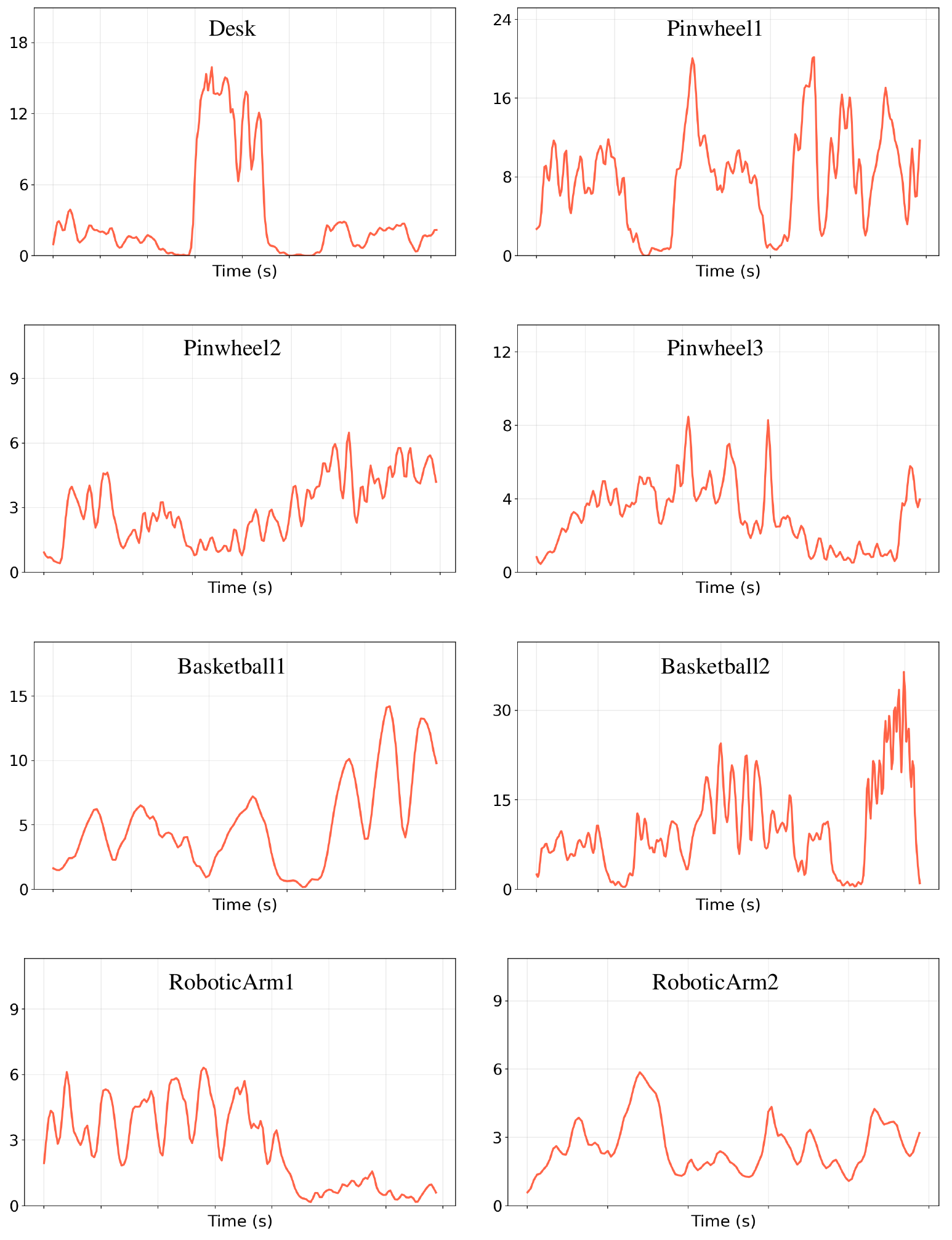}
    \vspace{-5pt}
    \caption{Average feature displacement over time for the DEFT dataset, illustrating the motion distribution.}
    \label{fig:det_distribution}
\end{figure*}

We introduce DEFT as an evaluation dataset that contains more challenging and highly dynamic motion patterns. For accurate annotations, we chose DAVIS346C hardware system, whose images and events are perfectly aligned. While several sequences were briefly presented in the main paper, we provide a more complete overview of the full dataset here. An overview of the dataset is given in Table~\ref{tab:det_overview}, the motion distribution is shown in Fig.~\ref{fig:det_distribution}, and representative scene examples are provided in Fig.~\ref{fig:det_sample}.

As noted in~\cite{messikommer2023data}, for the previously proposed EC and EDS datasets, a key limitation is that they mainly capture static scenes in which the camera moves around a fixed target. 
To address this limitation, DEFT includes more dynamic scenarios. For example, in the Basketball sequence, both the person and the camera move simultaneously, and in the RoboticArm sequence, the manipulator moves while the camera also moves in a different direction. These examples better reflect dynamic real-world tracking scenarios. In addition, the length of each sequence of DEFT is longer than conventional feature tracking evaluation dataset, which is designed to evaluate feature tracking under more deployment-oriented conditions. We compare the dataset length among the existing benchmarks and ours in Table~\ref{tab:dataset_length}.

\begin{table}[t]
\setlength{\tabcolsep}{10.2pt}
\centering
\caption{Comparison of the dataset length
}
\vspace{-8pt}
\label{tab:repre}
\renewcommand{\arraystretch}{1.1}
\resizebox{0.5\linewidth}{!}{
\begin{tabular}{@{}l|ccc@{}}
\hline
Dataset & EC & EDS & Ours \\
\hline
Total length (s) & 17.9 & 9.6 & 57.0 \\
Sequence \# & 5 & 4 & 8 \\
Min length (s) & 3.5 & 1.3 & 4 \\
Max length (s) & 3.7 & 4.0 & 10 \\
\hline
\end{tabular}
}
\label{tab:dataset_length}
\vspace{-12pt}
\end{table}

\section{More Experiments on Event Representation}
\label{sec:robust_event_repre}

We conduct experiments with various event representations to demonstrate that the proposed RL module operates in a representation-agnostic and general manner, rather than being dependent on a specific event representation. In addition to the event representations used in recent studies~\cite{shen2025blinktrack, messikommer2023data}, we further evaluate our method with two additional representations: event histogram~\cite{maqueda2018event} and voxel grid~\cite{zhu2019unsupervised}. Using the official implementation of BlinkTrack~\cite{shen2025blinktrack}, we train models with these different representations and then attach and train our RL module. The results are presented in Table~\ref{tab:repre}. The proposed method consistently improves performance across different event representations, demonstrating that it is agnostic to the event representation and can serve as a general approach.

\begin{table}[t]
\setlength{\tabcolsep}{10.2pt}
\centering
\caption{Performance on DEFT under different event representations.
}
\vspace{-6pt}
\label{tab:repre}
\renewcommand{\arraystretch}{1.1}
\resizebox{0.99\linewidth}{!}{
\begin{tabular}{l|cc|cc|cc}
\hline
\multirow{2}{*}{Method} & \multicolumn{2}{c|}{Event histogram} & \multicolumn{2}{c}{Voxel grid} & \multicolumn{2}{c}{SITS~\cite{manderscheid2019speed}}\\
 & FA$\uparrow$ & EFA$\uparrow$ & FA$\uparrow$ & EFA$\uparrow$ & FA$\uparrow$ & EFA$\uparrow$ \\
\hline
BT~\cite{shen2025blinktrack} & 0.153 & 0.144 & 0.144 & 0.132 & 0.284 & 0.143 \\
\textbf{BT}~\cite{shen2025blinktrack}\textbf{+Ours} & {0.446}(+0.313) & {0.439}(+0.295) & 0.425(+0.281) & 0.422(+0.290) & 0.555(+0.271) & 0.528(+0.385)\\
\hline
\end{tabular}
}
\label{tab:repre}
\vspace{-12pt}
\end{table}

\section{More Qualitative Results}
\label{sec:more_qual}

\begin{figure*}[p]
    \centering
    \includegraphics[width=0.99\linewidth]{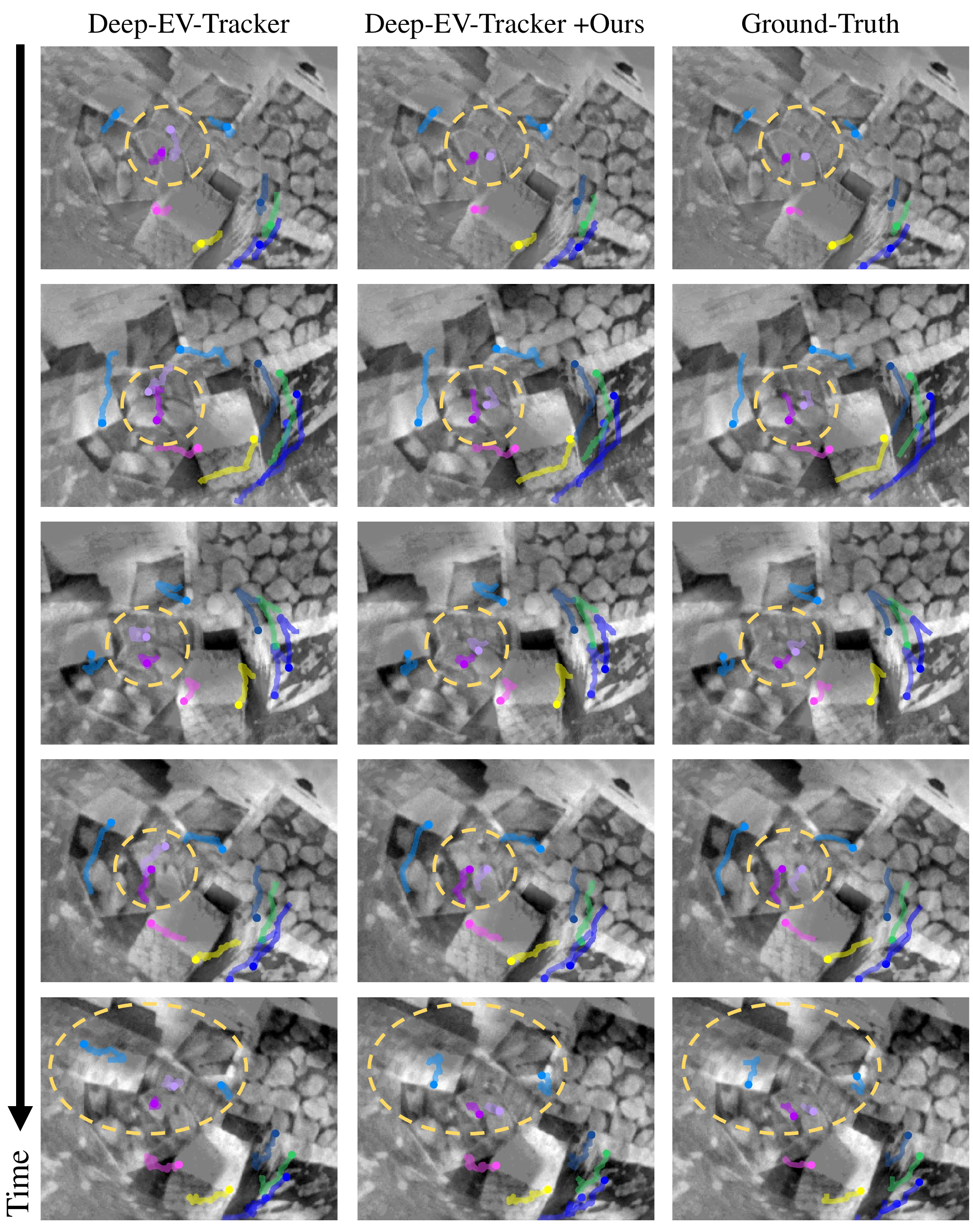}
    \vspace{-5pt}
    \caption{Qualitative performance comparison on the EC dataset. Events are accumulated over the interval for visualization, illustrating the temporal dynamics of the features during that period.}
    \label{fig:ec_qual}
\end{figure*}

\begin{figure*}[p]
    \centering
    \includegraphics[width=0.99\linewidth]{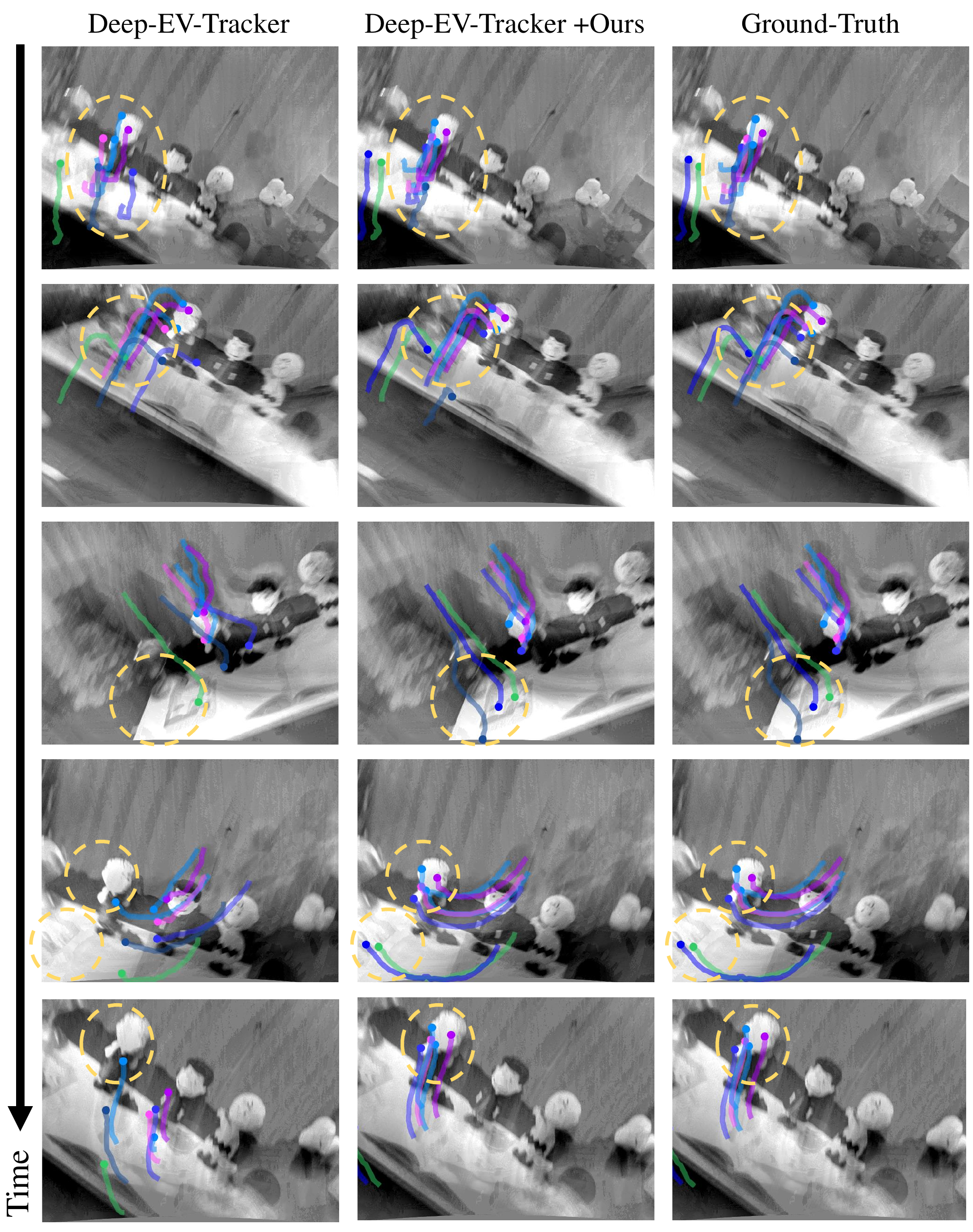}
    \vspace{-5pt}
    \caption{Qualitative performance comparison on the EDS dataset. Events are accumulated over the interval for visualization, illustrating the temporal dynamics of the features during that period.}
    \label{fig:eds_qual}
\end{figure*}

\begin{figure*}[p]
    \centering
    \includegraphics[width=0.99\linewidth]{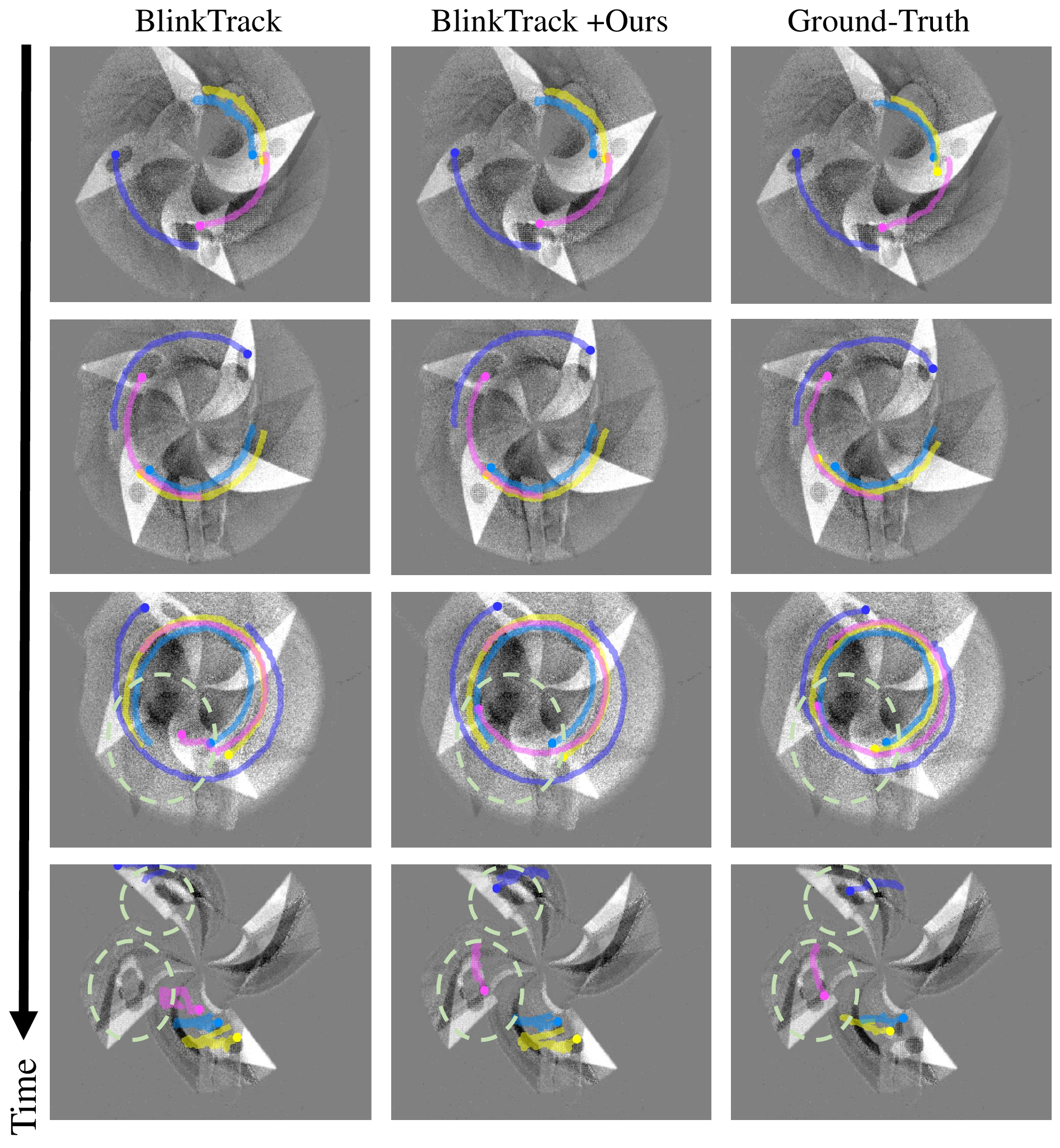}
    \vspace{-5pt}
    \caption{Qualitative performance comparison on the DEFT dataset. Events are accumulated over the interval for visualization, illustrating the temporal dynamics of the features during that period.}
    \label{fig:det_qual}
\end{figure*}

To demonstrate how the proposed method outperforms existing approaches, particularly in scenarios with highly dynamic motion distributions, we present qualitative results over time on the EC, EDS, and DEFT datasets in Figs.~\ref{fig:ec_qual}, \ref{fig:eds_qual}, \ref{fig:det_qual}.

\section{Behavior of the RL Module according to the Motion}
\label{sec:gostop}

\begin{figure*}[t]
    \centering
    \includegraphics[width=0.80\linewidth]{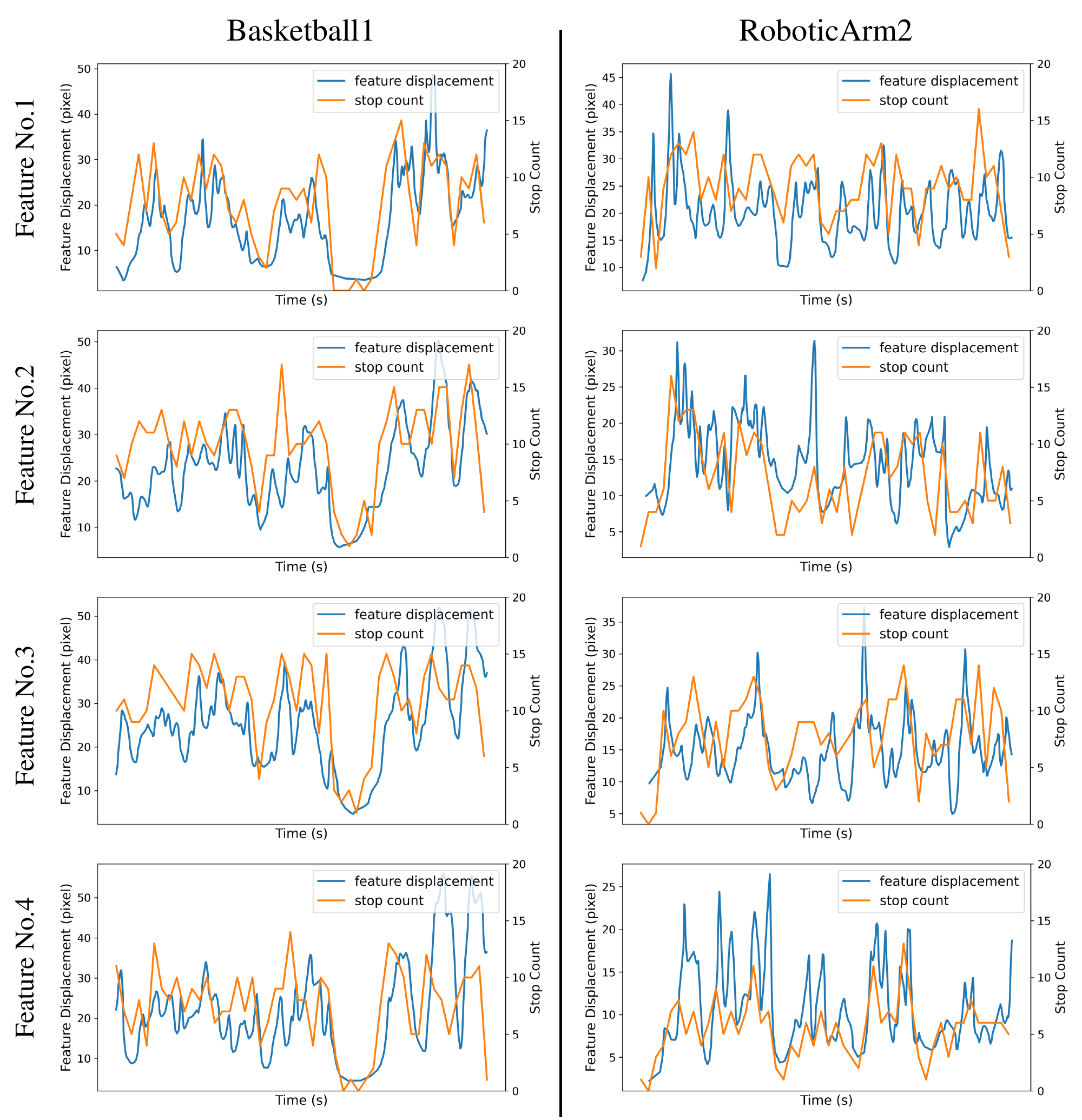}
    \vspace{-5pt}
    \caption{
    Visualization of the ground-truth feature displacement and the number of \textit{stop} actions performed by the proposed RL module over a 0.1-second interval. Each graph shows the statistics for a different feature. Larger displacement indicates faster motion, for which the RL module adaptively increases the \textit{stop} count. In contrast, when the displacement becomes smaller, indicating slower or nearly stationary motion, the \textit{stop} count decreases accordingly.
    }
    \label{fig:go_stop_qual}
    \vspace{-5pt}
\end{figure*}

\begin{figure*}[t]
    \centering
    \hspace{0.035\linewidth}
    \includegraphics[width=0.80\linewidth]{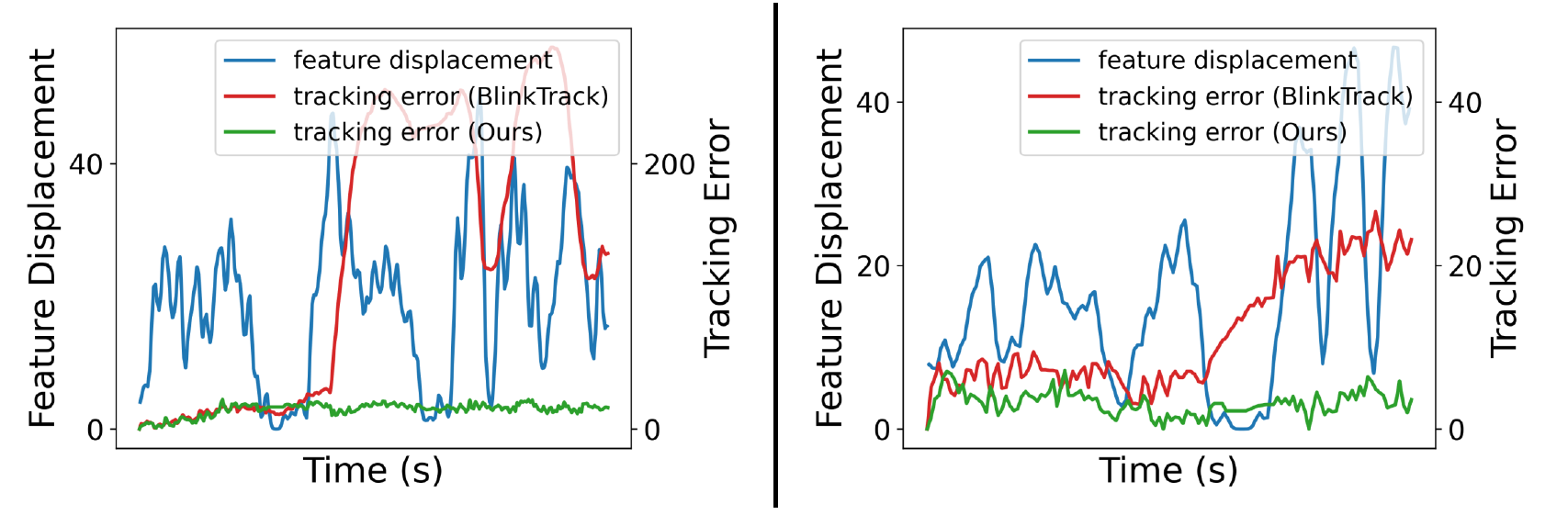}
    \vspace{-5pt}
    \caption{
    Visualization of feature displacement over time and the corresponding tracking error. The left and right graphs are taken from the Desk and Basketball1 sequences, respectively. The results show that baseline trackers often fail to maintain the target feature when its motion changes abruptly. In contrast, our method adaptively adjusts the temporal event window, enabling robust feature tracking.
    }
    \label{fig:go_stop_failure_cases}
\end{figure*}

Figure~\ref{fig:go_stop_qual} illustrates whether the proposed RL agent adaptively operates according to changes in the motion distribution. For each feature in the data, we present the feature displacement over a fixed interval together with the count of \textit{stop} actions output by the RL agent. The proposed RL agent increases the number of \textit{stop} actions as the motion becomes faster, enabling robust operation even under highly dynamic motion. In contrast, when the motion is small, the agent tends to select the \textit{go} action more frequently, achieving both computational efficiency and performance improvement even when event data is sparse. Figure~\ref{fig:go_stop_failure_cases} presents a case study illustrating when the proposed method improves tracking performance. While the baseline tracker loses the target feature under abrupt motion changes, our method adaptively adjusts the inference frequency based on the event stream input and enables robust feature tracking.

\begin{table*}[p]
\caption{An overview of the proposed DEFT dataset.}
\vspace{-6pt}
\centering
\setlength{\tabcolsep}{3.6pt}
\resizebox{1.0\textwidth}{!}{
\begin{tabular}{lcccc}
\thickhline
Name & Seconds & No. GT & No. Queries & Description \\
\hline
Desk & 4 & 100 & 5 & Various objects on a desk, moving dynamically with primarily translational motion. \\
Pinwheel1 & 10 & 250 & 3 & A pinwheel rotates at high speed with intermittent acceleration and deceleration. \\
Pinwheel2 & 8 & 200 & 10 & A pinwheel rotates slowly, with intermittent acceleration and deceleration. \\
Pinwheel3 & 8 & 200 & 5 & A pinwheel rotates while simultaneously moving along the z-axis. \\
Basketball1 & 5 & 125 & 5 &A basketball is balanced on one hand and moved in various directions. \\
Basketball2 & 10 & 250 & 5 & A basketball is held with both hands and rotated dynamically. \\
RobotArm1 & 7 & 175 & 5 & Balls are manipulated by a robot arm while the camera also moves dynamically. \\
RobotArm2 & 5 & 125 & 5 & Balls are moved by a robot arm while the camera undergoes dynamic motion. \\
\hline
\multicolumn{1}{l}{Total} & 57 & 1425 & 43  \\
\thickhline
\end{tabular}
}
\vspace{-20pt}
\label{tab:det_overview}
\end{table*}

\clearpage
%
%
\bibliographystyle{splncs04}
\bibliography{main}